\pdfoutput=1
 
\documentclass[11pt]{article}

\usepackage[preprint]{acl}

\usepackage{times}
\usepackage{latexsym}

\usepackage[T1]{fontenc}

\usepackage[utf8]{inputenc}

\usepackage{microtype}

\usepackage{inconsolata}

\usepackage{graphicx}

\usepackage{tikz}
\usepackage{subcaption}
\usepackage{pgfplots}
\usepackage{amsfonts,amssymb, amsthm, amsmath}
\usepackage{booktabs}
\usepackage{xcolor}
\usepackage[table]{xcolor} 

\usetikzlibrary{patterns,backgrounds}
\definecolor{c1}{RGB}{162, 0, 037}
\definecolor{c2}{RGB}{0, 87, 0}
\definecolor{c3}{RGB}{0, 110, 175}
\definecolor{c4}{RGB}{180, 101, 4}
\definecolor{c5}{RGB}{57,177,133}
\definecolor{RYB1}{RGB}{0,0,194}
\definecolor{RYB2}{RGB}{0,131,194}
\definecolor{RYB3}{RGB}{144,0,136}
\definecolor{RYB4}{RGB}{205,0,0}
\definecolor{tablecolor}{RGB}{218,232,252} 
\definecolor{light-gray}{gray}{0.95}
\definecolor{Gray}{gray}{0.9}

\definecolor{lightgreen}{RGB}{237,246,247}

\pgfplotsset{compat=1.18} 
\usepackage{lipsum} 
\usepackage{titlesec}
\usepackage{booktabs}
\usepackage{multirow}
\usepackage{bm}
\usepackage[normalem]{ulem}
\usepackage{colortbl}
\usepackage{epigraph}
\usepackage{paralist}
\newcommand{\paratitle}[1]{\vspace{1.2ex}\noindent \textbf{#1}}
\usepackage{enumitem}
\usepackage{pifont}
\usepackage{tcolorbox}
\tcbuselibrary{breakable}
\definecolor{qbox}{RGB}{172,172,172}
\definecolor{qinside}{RGB}{249,249,249}

\tcbuselibrary{listingsutf8}
\tcbset{
    mybox/.style={
        colback=qinside, 
        colframe=qbox, 
        coltitle=white, 
        fonttitle=\bfseries, 
        colbacktitle=qbox, 
        title={Explorations on CTEG}
    }
}

 \tcbset{
    mybox2/.style={
        colback=qinside, 
        colframe=qbox,
        coltitle=white, 
        fonttitle=\bfseries,
        colbacktitle=qbox, 
        title={Explorations on GiGA} 
    }
}
\usepackage{xspace}

\title{When Words Smile: Generating Diverse Emotional \\Facial Expressions from Text}
 
\author{Haidong Xu\textsuperscript{\rm 1}, Meishan Zhang\textsuperscript{\rm 1}, Hao Ju\textsuperscript{\rm 2}, Zhedong Zheng\textsuperscript{\rm 2} \\
\bf Erik Cambria\textsuperscript{\rm 3}, Min Zhang\textsuperscript{\rm 1}, Hao Fei\textsuperscript{\rm 4}\Thanks{ Corresponding author}  \\
\textsuperscript{\rm 1} Harbin Institute of Technology (Shenzhen), \, \textsuperscript{\rm 2} University of Macau \\
\textsuperscript{\rm 3} Nanyang Technological University, \, \textsuperscript{\rm 4} National University of Singapore\\
\texttt{182haidong@gmail.com, haofei37@nus.edu.sg}}

\begin{document}
\maketitle
\begin{abstract}
Enabling digital humans to express rich emotions has significant applications in dialogue systems, gaming, and other interactive scenarios. 
While recent advances in talking head synthesis have achieved impressive results in lip synchronization, they tend to overlook the rich and dynamic nature of facial expressions.
To fill this critical gap, we introduce an end-to-end text-to-expression model that explicitly focuses on emotional dynamics.
Our model learns expressive facial variations in a continuous latent space and generates expressions that are diverse, fluid, and emotionally coherent.
To support this task, we introduce EmoAva, a large-scale and high-quality dataset containing 15,000 text–3D expression pairs. 
Extensive experiments on both existing datasets and EmoAva demonstrate that our method significantly outperforms baselines across multiple evaluation metrics,
marking a significant advancement in the field.
\footnote{Resources are available at this \href{https://WalkerMitty.github.io/EmoAva}{link}.}
\end{abstract}

\section{Introduction}
\label{intro}

In recent years, the remarkable success of dialogue systems has sparked a growing desire for face-to-face interaction with digital humans~\cite{dialogue1}.
As emotional beings, humans rely heavily on facial expressions as a primary means of conveying emotions and intentions. 
Therefore, enabling digital humans to express emotions through facial expressions holds substantial research and application value~\cite{laughtalk}.

A large portion of digital human (also referred to as talking head) research~\cite{TalkingGaussian,emotalk3D}, focuses primarily on the synchronization between speech and lip movements, while largely ignoring the rich emotional and expressive dynamics of face-to-face communication. 
Although some previous studies have recognized this limitation and started investigating potential solutions, most existing systems~\cite{emotalk3D,EMOTE,expressiveTalking} still generate only coarse facial expressions based on a limited set of discrete emotion labels (e.g., “happy”, “sad”), usually employing a pipeline architecture.

\begin{figure}[!t]
  \centering
  \includegraphics[width=\columnwidth]{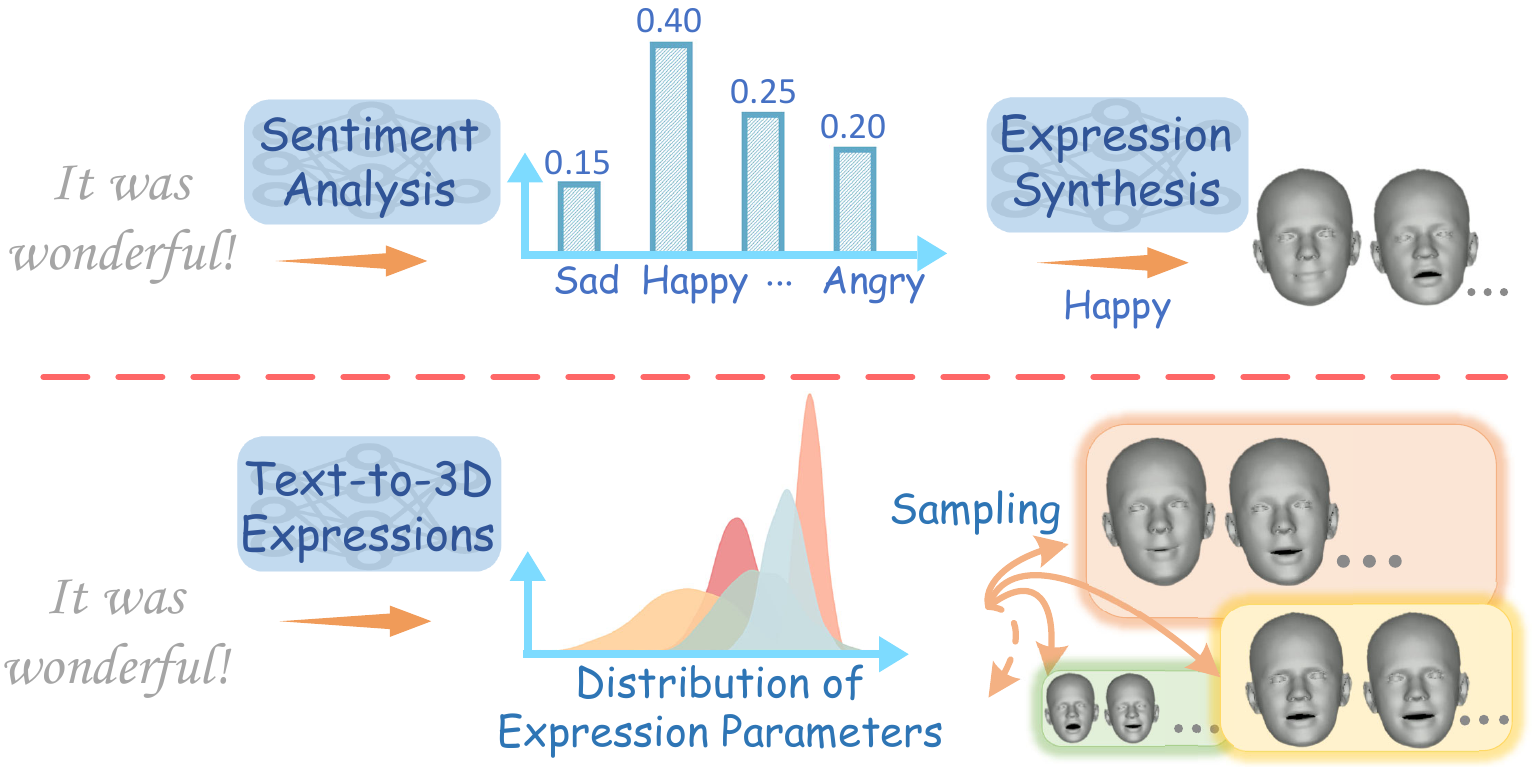}
\caption{
    \textbf{Top:} The existing pipeline for synthesizing emotional avatars, which can only generate limited expressions that lack of diversity.
    \textbf{Bottom:} The proposed end-to-end system that directly maps text to facial expressions (codes), aims to generate diverse, emotionally consistent, and temporally smooth expressions.
}

  \label{fig:intro}
\vspace{-3mm}
\end{figure}

Following the current pipeline, one typically performs sentiment analysis on the speech or text to obtain a discrete emotion label, which is then used to condition the facial expression generation (as shown in Figure~\ref{fig:intro}).
However, this approach faces at least two major limitations. First, emotion labels are typically limited in number and struggle to capture the full richness and subtlety of human emotional expression. 
Second, pipeline-based models are susceptible to information loss and error propagation across stages.
End-to-end modeling could be one promising strategy to address the above limitations.
By generating a continuous sequence of 3D facial expressions directly from the input utterance,
the output is expected to be more diverse, natural, and emotionally consistent.

In this work, we present the first end-to-end work for text-3D facial expression learning.
Technically, we propose a unified model, \textbf{CTEG} (Continuous Text-to-Expression Generator).
CTEG leverages a CVAE-based autoregressive architecture to model expressive variations \textbf{in a continuous latent space}, enabling smooth and natural expression synthesis. 
To ensure emotional consistency between the input text and generated expressions, CTEG adopts a Latent Temporal Attention (LTA) mechanism that enhances the latent representation at each timestep by attending to historical context. 
Additionally, to promote expressive richness, CTEG incorporates an Expression-wise Attention (EwA) module that captures spatial dependencies among facial regions, enabling coordinated and varied facial movements.

To facilitate end-to-end training, we introduce \textbf{EmoAva}, a high-quality, large-scale dataset comprising 15,000 text-to-3D expression mapping instances, collected from multi-party dialogue scenes in professionally acted video sources. EmoAva provides rich, emotionally diverse, and context-aware expressive behaviors, offering a valuable foundation for studying facial expression generation in conversational AI.
Extensive qualitative and quantitative experiments on the EmoAva as well as other representative existing datasets~\cite{lm_listener}
demonstrate the superiority of our CTEG model in terms of expression diversity, naturalness, and emotional consistency, establishing a strong baseline for future research in text-to-expression generation.

In summary, our key contributions are as below.
\begin{itemize}[topsep=0pt,noitemsep]
    \item We propose a novel end-to-end model, CTEG, which learns text-to-expression mapping in a continuous latent space.
    \item We introduce \textbf{EmoAva}, a high-quality dataset with 15,000 annotated instances, designed to alleviate data scarcity in this domain.
    \item Extensive experiments demonstrate the effectiveness of CTEG in capturing expression diversity, naturalness, and emotional consistency, establishing a strong baseline for future research in this field.
\end{itemize}

\section{Related work}

\paragraph{Speech-driven Emotional Avatar Synthesis.}

Extensive research has been conducted on the synthesis of 3D talking heads~\cite{meshtalk,2d-avatar-1,2d-avatar-2,faceformer,face2face,sadtalker,osm-net}, most of which are speech-driven—generating lip-synced facial animations from audio input. 
These works generally overlook the modeling of facial expressions. 

Recently, some approaches have started to integrate emotional context into the generation process~\cite{emotalk3D}.
EMOTE~\cite{EMOTE} addresses this by controlling expressions with single emotional labels, but the limited categories do not capture the full range of human emotions. 
Conversely, EmoTalk~\cite{emotalk} and LaughTalk~\cite{laughtalk} extract tonal emotional features from speech to guide avatar synthesis similarly to talking head tasks.
Complementary to these approaches, our method explores text as the sole source of emotional input since textual dialogue inherently conveys rich affective information and is more abundantly available than other modalities~\cite{sentiment_analysis1}.

\paragraph{Text-Driven Human Motion Generation.}
Text-based human motion generation has significant applications in areas such as gaming and virtual reality. 
Much of the existing research in this domain focuses on generating sequences of human body movements~\cite{T2M-GPT,motionGPT}. 
In contrast, relatively little attention has been paid to text-driven human facial expression generation~\cite{lm_listener,listener2}.

Our approach is most closely related to the recent work LM-Listener~\cite{lm_listener}, which focuses on generating listener motions. 
In contrast, we concentrate on the more diverse and complex expressions of speakers.
While LM-Listener employs a VQ-VAE-based framework, the use of discrete modeling and a constrained latent space may limit its ability to maintain temporal coherence and capture the full range of speaker-driven facial dynamics.
In comparison, our method adopts a CVAE framework, whose structured continuous latent space is better suited for modeling the fluidity and expressiveness of facial behaviors.

\section{Our Approach}
\label{CTEG}
\subsection{Task Definition}
Given a text input $\mathbf{x}$, our system generates a sequence of expression vectors $\psi= \{\mathbf{E}_{0},\mathbf{E}_{1}, ... , \mathbf{E}_{T} \}$ 
over $T$ time steps. 
The expression vectors are derived from 3D Morphable Face Model (3DMM) frameworks~\cite{FaceWarehouse,3DMM,lm_listener}, which parameterize facial geometry in a compact and interpretable form. Among various 3DMMs, we follow~\citet{lm_listener} and adopt the widely used FLAME model~\cite{FLAME}, denoted as $\mathcal{F}$.
Specifically, FLAME defines the parameter set $\mathcal{F} = \{\beta, \Delta v, \varrho, \Theta, \psi\}$, where $\beta$ denotes shape, $\Delta v$ represents vertex offsets, $\varrho$ is global translation, $\Theta$ denotes joint poses, and $\psi$ captures expression-related deformations.
FLAME decouples expression from the shape and other identity-related factors, allowing us to directly regress the expression parameters $\psi \in \mathbb{R}^d$ in an identity-agnostic manner, where $d$ (53 in this paper) denotes the dimension of the expression space.
\subsection{Overall framework}
\begin{figure}[!t]
  \centering
  \includegraphics[width=\columnwidth]{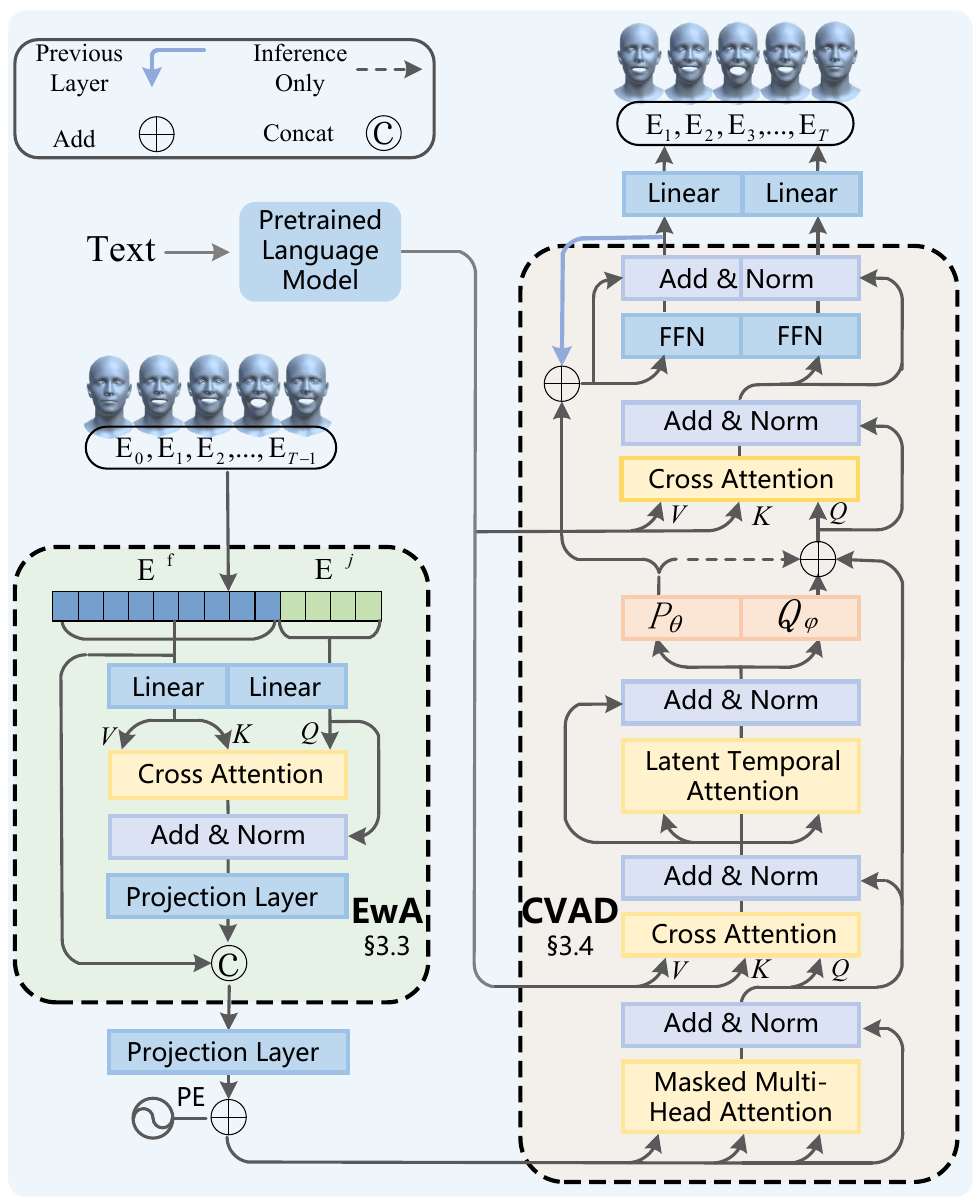}
  \caption{
  Architecture of the Continuous Text-to-Expression Generator (CTEG). Given a text, the model autoregressively generates a sequence of expression vectors. The
  \colorbox[rgb]{0.924, 0.956, 0.902}{green block} and \colorbox[rgb]{0.992, 0.957, 0.933}{pink block} represent the proposed Expression-wise Attention (EwA) module and the core Conditional Variational Autoregressive Decoder (CVAD) module, respectively. 
  }
  \label{fig:T3E}
\vspace{-3mm}
\end{figure}

The overview framework of the CTEG model is shown in Figure \ref{fig:T3E}. 
CTEG primarily consists of the Expression-wise Attention (EwA) block at the encoder side, and the Conditional Variational Autoregressive Decoder (CVAD) block at the decoder side.
From the perspective of architectural design, the EwA module serves as a feature enhancement module, while the CVAD functions as a hybrid of a CVAE~\cite{cvae} and a transformer decoder~\cite{transformer}.
Technically, we adopt such an architecture for the following advantages.
\textbf{1)} CVAE is beneficial for maintaining a smooth spatial distribution due to its nature of modeling in continuous space~\cite{vae,cvae}, which may help to model expression fluidity.
\textbf{2)} Transformer decoder excels at modeling the long-range dependencies between sequences~\cite{transformer}, which may help to model the emotion-content consistency.
\textbf{3)} Due to the richness of the facial expression sequence, even within a single time step, facial expressions have countless variations.
The Variational Autoregressive Decoder (VAD) may facilitate the modeling of diverse, time-varying sequences~\cite{VAD}.
Given a text $\mathbf{x}$ as input, CTEG generates a sequence of expression vectors $\psi$ autoregressively.

\vspace{-2mm}
\subsection{Expression-wise Attention Module}
\label{EwA}
In the input part, we introduce EwA to establish connections between facial units and enhance the richness of the input expression in the feature space.
This guides the subsequent CVAD module to capture different and rich patterns and structures, thereby improving the overall diversity of the model's generation results.

The expression vector $\mathbf{E}$ is constructed by concatenating two components: the jaw part $\mathbf{E}^j$ and the above-jaw part $\mathbf{E}^f$.
Intuitively, these two parts are not independent of each other because human facial units function as a whole. 
For example, when a person laughs heartily, the jaw controls the opening of the mouth.
To establish a connection between them, we first use a projection layer to map the raw expression vector to a latent space.
Then we let the transfered $\mathbf{E}^j$ as the query, and let the transfered $\mathbf{E}^f$ as the key and value, feeding them into a cross attention module~\cite{transformer}. 
After that, we apply dimensionality reduction to the output of the attention module and obtain $\mathbf{E}^{j'} \in \mathbb{R}^{|\mathbf{E} ^j|}$.
The final recombined 3D expression codes $\mathbf{E'} \in \mathbb{R}^{|\mathbf{E} |}$ are represented by:
$\mathbf{E'} = \text{Concat}(\mathbf{E}^f, \mathbf{E}^j+\mathbf{E}^{j'})$.
Then we project it into a high dimension $d_{model}$.
In order to capture the order of expression sequence, we add the Positional Embeddings (PE) to the output of the EwA module.
Specifically, we adopt the sinusoidal version positional encodings introduced in~\citet{transformer}.

\vspace{-3mm}
\subsection{Conditional Variational Autoregressive Decoder}

\vspace{-2mm}

Given a text $\mathbf{x}$ as input, an expression sequence $\psi$ as output, CVAE is to maximize the conditional log-likelihood $\log{p(\psi|\mathbf{x})}$. 
To better capture temporal dynamics, we model the conditional probability distribution at each time step.
Formally, the log-likelihood in our method is $\log{\prod_{t=1}^{T} p(\psi_{t} \mid \psi_{<t},\mathbf{x})}$, rather than $\log{p(\psi_{0,...,T} \mid \mathbf{x})}$.
To enhance the emotion-content consistency, we explicitly model the historical states of the latent variables.
The resulting generation model can be formulated as:
\setlength{\abovedisplayskip}{4pt}
\setlength{\belowdisplayskip}{4pt}
\setlength{\abovedisplayshortskip}{4pt}
\setlength{\belowdisplayshortskip}{4pt}
\begin{equation}
\begin{aligned}
p(\psi \mid \mathbf{z},\mathbf{x}) &= \textstyle\prod_{t=1}^{T} p(\psi_{t} \mid \psi_{<t}, \mathbf{z}_{t},\mathbf{x}) \\
&= \textstyle\prod_{t=1}^{T} p(\psi_{t} \mid \psi_{<t}, f_{\zeta}(\mathbf{z}_{<t}), \mathbf{x}) \,,
\end{aligned}
\end{equation}
where $f_{\zeta}$ is the Latent Temporal Attention (LTA) module, implemented by the masked multi-head attention~\cite{transformer}.
\footnote{We also try another simple model, the details can be found in the Appendix~\ref{appendix_cteg}}.

Intuitively, we assume the prior distribution $P_{\theta}$ and conditional distribution $Q_{\phi}$ to be multivariate Gaussian distributions:
\begin{equation}
\begin{aligned}
&Q_{\phi}(\mathbf{z}_{t} \mid \psi_{\le t}, \mathbf{x}) = \mathcal{N}(\boldsymbol{\mu}
_{r}(\psi_{\le t}, \mathbf{x}),\boldsymbol{\sigma}_{r}(\psi_{\le t}, \mathbf{x})) \,, \\
&P_{\theta}(\mathbf{z}_{t} \mid \psi_{<t}, \mathbf{x}) = \mathcal{N}(\boldsymbol{\mu}
_{p}(\psi_{< t}, \mathbf{x}),\boldsymbol{\sigma}_{p}(\psi_{< t}, \mathbf{x})) \,.
\end{aligned}
\end{equation}
The two Gaussian distributions are parameterized by two neural networks respectively:
\begin{equation}
\begin{aligned}
&[\boldsymbol\mu_{r},\boldsymbol\sigma_{r}] = 
[h_{r}^{\mu}(\textbf{o}), h_{r}^{\sigma}(\textbf{o})] \,, \\
&[\boldsymbol\mu_{p},\boldsymbol\sigma_{p}] = 
[h_{p}^{\mu}(\textbf{o}),h_{p}^{\sigma}(\textbf{o})] \,, \\
&\textbf{o} = \mathcal{A}_{mask}[\mathcal{A}(\psi_{\le t}, \mathbf{x}) \,,
\end{aligned}
\end{equation}
where $h$ denotes a linear layer, $\mathcal{A}_{mask}$ and $\mathcal{A}$ denote masked attention module and cross attention module, respectively.
As sampling $\mathbf{z}_{t}$ from two distributions is non-differentiable, we employ the reparameterization trick~\cite{vae}: 
\begin{equation}
\mathbf{z}_{t} = \mathrm{\mu}_{t}+ \mathrm{\sigma}_{t} \odot \epsilon, \epsilon \sim \mathcal{N}(0,\mathrm{I}) \,,
\end{equation}
$\mathbf{z}_{t}$ is drawn from $Q_{\phi}(\mathbf{z}_{t} \mid \psi_{\le t}, \mathbf{x})$ in the training stage, while drawn from $P_{\theta}(\mathbf{z}_{t} \mid \psi_{<t}, \mathbf{x})$ in the inference stage. 
After we obtain the sampled $\mathbf{z}_{t}$, we learn the second conditional generation distribution $P_{\theta}(\psi_{t} \mid \psi_{<t}, \mathbf{z}_{t}, \mathbf{x}) $.
Similarly, we assume the distribution a multivariate Gaussian distribution, the mean $\mu_{g}$ can be parameterized by the following generation network:
\begin{equation}
\begin{aligned}
&\mathbf{\mu}_{g}^t=\mathrm{FFN}(\mathrm{Concat}(\mathcal{A}(\textbf{o}_{1}),\mathcal{A}(\textbf{o}_{2}),...,\mathcal{A}(\textbf{o}_{l}))) \,,\\
&\mathcal{A}(\textbf{o}_{i}) = \mathcal{A}((\psi_{<t}+ \mathbf{z}_{<t})W^Q_{i}, \mathbf{x}W^K_{i}, \mathbf{x}W^V_{i}) \,,
\end{aligned}
\end{equation}
where $l$ is the number of cross attention heads.
$\text{FFN}$ is the position-wise feed-forward network.

Note that the CVAD module can be stacked multiple layers deep, where the input of the first layer comes from the EwA module, and the input of each subsequent layer comes from $\mathbf{\mu}_{g}$ obtained by the previous layer.
Specifically, the input at layer $m$ is sampled from $\mathcal{N}(\mu_{g}^{m-1},\sigma)$. 
For simplicity, we parameterize only $\mu_{g}$ and set the $\sigma$ of the generative distribution to a matrix where all entries are equal to $1$.
Finally, we sample the predicted expression codes in time step $t$ using the reparameterization trick again:
$\hat{\psi}_{t} = \mu_{g}^{m,t} + \epsilon , \epsilon \sim \mathcal{N} (0,\mathrm{I})$.
Our loss function of CVAD is as follows:
\setlength{\abovedisplayskip}{4pt}
\setlength{\belowdisplayskip}{4pt}
\setlength{\abovedisplayshortskip}{4pt}
\setlength{\belowdisplayshortskip}{4pt}
\begin{equation}
\label{loss_cvad}
  \mathcal{L}_{CVAD} = \sum_{t}^{}\mathcal{L}_{rec}(\psi_{t},\hat{\psi}_{t}) +\sum_{t}^{} \mathcal{L}_{KL}(t) \,,
\end{equation}
where $\mathcal{L}_{rec}$ is the mean squared error (MSE) loss, and the corresponding Kullback-Leibler (KL) divergence term is defined as follows:
\setlength{\abovedisplayskip}{0pt}
\setlength{\belowdisplayskip}{0pt}
\setlength{\abovedisplayshortskip}{0pt}
\setlength{\belowdisplayshortskip}{0pt}
\begin{align}
  \mathcal{L}_{KL}(t) = \mathrm{KL}\Big(& Q_{\phi}(\mathbf{z}_{t} \mid \psi_{\le t}, \mathbf{x}) \notag \\
  &\parallel\; P_{\theta}(\mathbf{z}_{t} \mid \psi_{<t}, \mathbf{x}) \Big) \,.
\end{align}

\vspace{-2mm}
\subsection{Target Guided Loss}
\label{target_loss}
Despite the excellent performance, CVAD is known to suffer from a notorious issue referred to as \textit{model collapse}~\cite{cyclical,yang2017}. 
When this happens, the $\mathrm{KL}$ divergence term in the loss function becomes very close to zero, and the latent variable may be ignored by the decoder.

Some works have proposed various methods to mitigate this issue~\cite{cyclical,yang2017,VAD}. 
Among them, the most common approach is to adjust the weight of the $\mathrm{KL}$ term during the training process.
However, this method requires carefully selecting parameters based on the specific training process of models, which is time-consuming when the dataset is very large or the model is very large.
To this end, we design a simple yet effective loss function $\mathcal{L}_{g}$ to guide the latent variables to learn a meaningful structure.
In this way, the latent variables may guide the autoregressive model in its generation process, thereby preventing the model from directly ignoring the latent variables.
Formally,
\setlength{\abovedisplayskip}{0pt}
\setlength{\belowdisplayskip}{4pt}
\setlength{\abovedisplayshortskip}{0pt}
\setlength{\belowdisplayshortskip}{4pt}
\begin{equation}
\label{eq:target_loss}
\begin{aligned}
&\mathcal{L}_{g} = \sum_{t}^{} \mathcal{L}_{rec} (\psi_{t},f_{\gamma }(\mathbf{o}_{t})) \,, \\
&\mathbf{o}_{t}=\sum_{i=1}^{N_{c}}\mathrm{FFN}_{i}(\mathbf{z}^i_{<t}) \,,
\end{aligned}
\end{equation}
where $f_{\gamma}$ is a linear projection layer, $N_{c}$ is the number of the CVAD layers.

\vspace{-2mm}
\subsection{Details of Training and Inference}
\label{CTEG_settings}
The total loss function of CTEG is: 
\begin{equation}
  \mathcal{L}_{total} = \mathcal{L}_{CVAD}+ \mathcal{L}_{g} \,,
  \label{CTEG_loss}
\end{equation}
We freeze the pretrained language model parameters during training and update the remaining parameters using the backpropagation algorithm.
We use the teacher forcing~\cite{teacher-forcing} approach to train in parallel for speed, but decode the expression codes sequentially during the inference phase.
Adam~\cite{adam} optimizer is adopted here.
We train $100$ epochs with the maximum expression sequence length $256$ and maximum sentence length $128$.
We set the warm-up steps $\textit{warmup}=4000$ and adopt the same learning rate scheduler in~\citet{transformer}:
\begin{equation}
  \textit{lr} = d_{model}^{-0.5} \cdot \min(\textit{step}^{-0.5},\textit{step} \cdot \textit{warmup}^{-1.5}) \,.
\end{equation}

Additionally, we also employ residual connection \cite{he2016deep} and layer normalization~\cite{layernorm} throughout our architecture, as detailed in Figure \ref{fig:T3E}.
The number of heads in all our attention modules is set to $12$.
The inner hidden dimension for the FFN model is $2048$.
Following~\citet{motion_use_bert}, we adopt the pretrained language model BERT\cite{bert} to obtain meaningful sentence embeddings. 
Specifically, the last hidden state of $\textit{bert-base-cased}$ is used here.
Both the text embedding and the expression embedding share the same feature dimension $d_{model} = 768$.
We only use one single CVAD layer in this paper (i.e., $N_{c}=1$).
More details about model analysis can be found in the Appendix~\ref{appendix_cteg}.

To ensure smoothness in the predicted sequences, an \emph{average} expression—computed as the mean of all expression parameters in the training set—is prepended to each sequence.
We treat an expression with all parameters set to zero as a terminator, referred to as the ``\emph{standard}'' face. 
Since sampling in a continuous space differs from discrete space sampling in language models, optimizing a single point in continuous space as a terminator is more challenging. 
One straightforward approach involves setting a threshold (e.g., $1.0$) during the inference phase, stopping when the predicted expression is very close to the "standard" face based on the Euclidean distance. 
We further employ a length-constrained decoding method by setting a maximum sequence length (MSL).

\begin{figure}[!t]
  \centering
  \includegraphics[width=0.9\columnwidth]{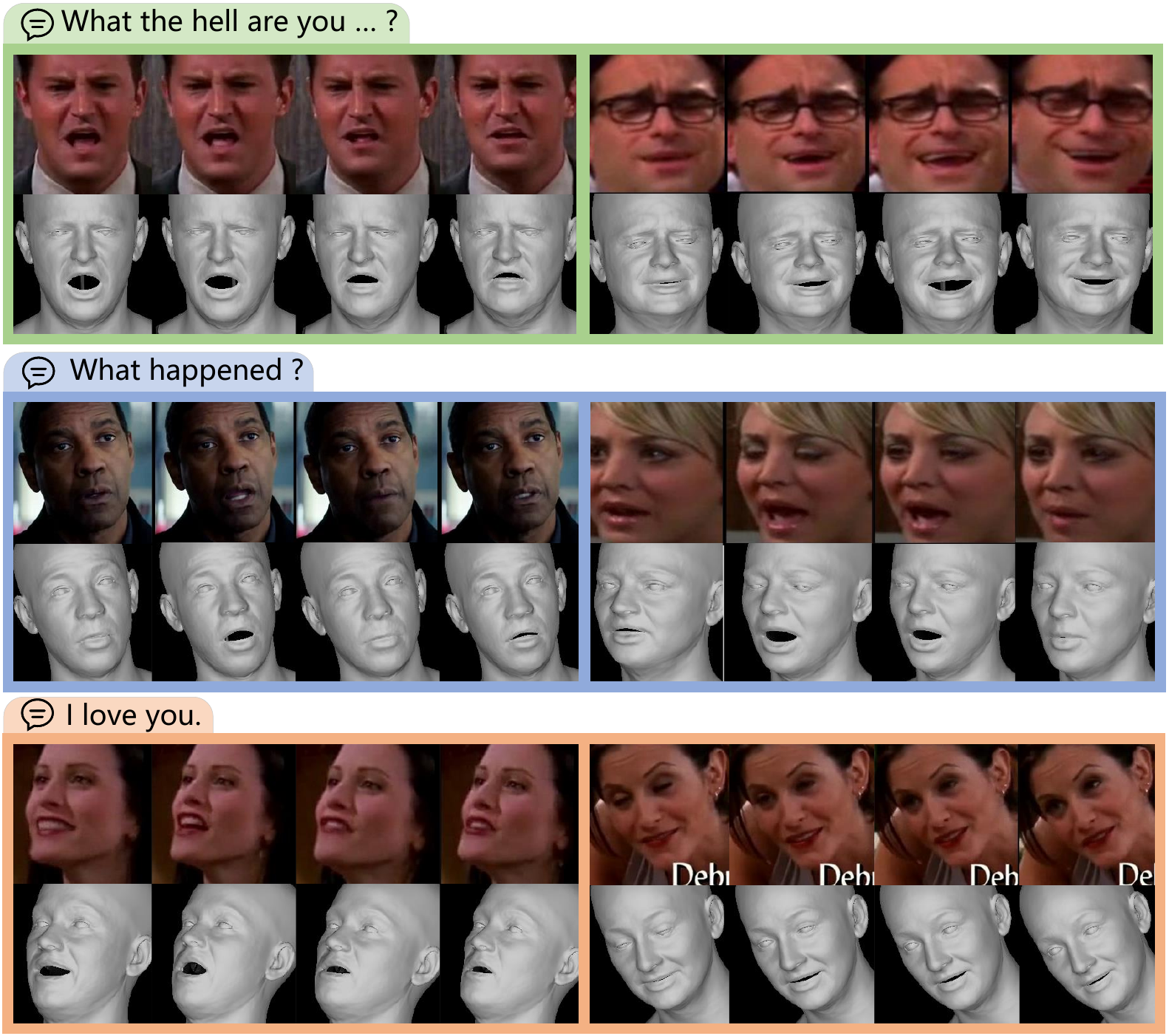}
  \caption{Samples from \textbf{EmoAva} dataset.  
  Each instance includes a textual dialogue spoken by an actor, a corresponding head video, and a sequence of 3D expression vectors (here visualized in 3D mesh).}
  \label{fig:dataset-samples}
\vspace{-2mm}
\end{figure}

\begin{table}[t!]
\fontsize{8}{8.5}\selectfont
\setlength{\tabcolsep}{1.5mm}
\centering

\begin{tabular}{lccccc}
\toprule
Dataset & \#Train & \#Validation & \#Test &Multi-person? \\
\midrule
\citet{lm_listener} & 2,366 & 222 & 543 & \ding{55}\\
EmoAva & 12,000 & 1,500 & 1,500 &\ding{51}  \\
\bottomrule
\end{tabular}

\caption{Comparisons of text-to-3D expression datasets. EmoAva is significantly larger and more diverse, featuring characters from over 100 screen productions.}
\label{tab:comparison}
\vspace{-4mm}
\end{table}

\section{EmoAva Dataset} 

We expect each instance in EmoAva to include a piece of text to be spoken, and a corresponding sequence of 3D expression vectors, as illustrated in Figure \ref{fig:dataset-samples}.
To construct this data, we first gather a large number of video clips from TV series and movies with dialogues.
We consider two existing data sources for multimodal emotion analysis task, MELD~\cite{MELD} and MEMOR~\cite{MEMOR}, both of which consist of television show segments.
Besides these, we also gather numerous video clips from YouTube.
A total of 21,390 such raw clips are collected, all in English.

We apply various preprocessing methods. A brief overview is provided below, with details in the Appendix~\ref{appendix_dataset}.
We employ WhisperX~\cite{whisperX} to transcribe the audio, resulting in the corresponding text and timestamps.
Afterwards, we cut the videos via the timestamps, creating dialogue video segments corresponding to the texts. 
To obtain clean headshot segments for each speaker, we develop a two-stage speaker localization pipeline. Specifically, we first apply FaceNet~\cite{facenet} for automatic face tracking. 
To handle the challenges posed by complex visual scenes—such as multiple characters in a single frame or frequent speaker switches—we further perform manual refinement to ensure accurate speaker segmentation.
After obtaining the head-videos, we adopt a 3D face tracking model EMOCA-v2~\cite{emoca} to extract the 3D expressions from 2D videos.

We collect a total of 15,000 text-to-expression pairs. 
A comparison with existing dataset is shown in Table~\ref{tab:comparison}. 
The dataset contains 782,471 FLAME frames. Among the 15,000 pairs, 2,270 exhibit a one-to-many (1-to-N) relationship—where N ranges from 2 to 76.

\vspace{-2mm}

\section{Experimental Setup}
\subsection{Evaluation Methods}

As introduced in Section~\ref{intro}, our system aims to generate a sequence of expressions that are diverse, fluid, and consistent with the conveyed emotional content.
To evaluate this, we adopt several evaluation metrics from prior studies and
introduce additional fine-grained evaluation criteria.
The calculation formulas for the following metrics are presented in the Appendix~\ref{appendix_metrics}.

\vspace{-2mm}
\paragraph{Diversity} measures the diversity of generated sequences without text as conditions~\cite{T2M-GPT}. 
We randomly sample $N/2$ sequence pairs and calculate the average Euclidean distance between the expression vectors.
\vspace{-2mm}
\paragraph{Multimodality} (abbreviated as MModality) measures the diversity of text-conditioned results~\cite{T2M-GPT}. 
We generate two expression sequences per text and compute their average Euclidean distance.
\vspace{-2mm}
\paragraph{Variation} measures the diversity of a sequence as it changes over time~\cite{lm_listener}.
\vspace{-2mm}
\paragraph{Fine-grained Diversity} (abbreviated as FgD) quantifies the subtle temporal fluctuations within facial expression sequences that are not fully captured by existing diversity metrics like Variation~\cite{lm_listener}. While Variation measures overall sequence diversity over time, FgD focuses specifically on the average Euclidean distance between adjacent frames to capture rapid, fine-grained changes in expressions.

\vspace{-2mm}

\paragraph{Diversity on Test} (abbreviated as DoT) measures the diversity of the expression sequences generated from the test set texts from a macro perspective.

\vspace{-2mm}

\paragraph{Continuous perplexity}(abbreviated as Cppl) evaluates how naturally an expression sequence evolves over time, reflecting the smoothness of the expression sequence and the uncertainty in a model's predictions. 
Contrary to the traditional discrete perplexity metric~\cite{perplexity}, Cppl is computed in continuous space.

\vspace{-2mm}
\label{consistency}
\paragraph{Consistency} assesses the extent to which the expressions accurately represent the emotions that would naturally correspond to a given utterance.
Due to the absence of precise automatic tools to evaluate this alignment, we rely on human evaluation following \citet{T2M-GPT}.

\begin{table*}[!t]

\centering
\scriptsize
\resizebox{0.95\linewidth}{!}{%
\begin{tabular}{lllllll}
\toprule
    & Cppl $\downarrow$      & DoT $\rightarrow$      & FgD $\rightarrow$      & Diversity $\uparrow$  & MModality $\uparrow$  & Variation $\rightarrow$  \\ \cmidrule{2-7}
\rowcolor{light-gray}
\textit{GT}   & /       & $9.24$     & $1.26$     & /      & /      & $0.28$     \\
Shuffle   & $9.90e6$       & /     & /     & /      & /      & /     \\
LM-Listener~\cite{lm_listener}  & /       & $7.99$     & $1.04 \pm 0.0039$     & $7.01 \pm 0.0736$     & $6.40 \pm 0.0662$     & $0.32 \pm 0.0048$     \\
\rowcolor{Gray}
CTEG ($MSL=256$)  & $\bm{{262.19}}$     &$\mathbf{9.96}$ & $\bm{{1.22}} \pm 0.0020$ &$\bm{{8.18}} \pm 0.0508$ & $\bm{{9.35}} \pm 0.0430$ & $0.72 \pm 0.0055$ \\ 
\rowcolor{Gray}
CTEG ($MSL=64$) & / & $8.91$     & $1.22 \pm 0.0021$     & $7.75 \pm 0.0535$     & $8.48 \pm 0.0485$     &$\bm{{0.31}} \pm 0.0059$     \\

\bottomrule
\end{tabular}%
}
\caption{Main Quantitative results. CTEG significantly outperforms the LM-Listener across four diversity metrics and achieves notably lower perplexity compared with the \textit{Shuffle} setting. 
$\downarrow$ indicates that a lower value is better while $\uparrow$ suggests that a higher value is preferable. 
$\rightarrow$ indicates that the closer the value is to the \textit{GT}, the better.
$/$ indicates not applicable.
The standard error is estimated through bootstrap resampling with $1,000$ iterations.
}

\label{CTEG_quantitative_all}
\end{table*}

\begin{table*}[!t]

\centering
\footnotesize
\resizebox{0.98\linewidth}{!}{%
\tiny
\begin{tabular}{lllllll}
\toprule
    & Cppl $\downarrow$      & DoT $\rightarrow$      & FgD $\rightarrow$      & Diversity $\uparrow$  & MModality $\uparrow$  & Variation $\rightarrow$  \\ \cmidrule{2-7}
\rowcolor{light-gray}
\textit{GT} & / & $9.24$     &$1.26$     & /     & /     & $0.28$     \\
w.o. EwA & $\bm{{205.65}}$ & $6.97$     & $1.14 \pm 0.0018$     & $6.07 \pm 0.0246$     & $6.55 \pm 0.0293$     & $0.40 \pm 0.0025$     \\
w.o. LTA & $243.57$ & $8.60$
    & $1.20 \pm 0.0030$
 & $7.58 \pm 0.0484$
 & $8.01 \pm 0.0671$
 & $0.58 \pm 0.0062$
     \\
w.o. $\mathcal{L}_{g}$ & $646.34$ & $7.67$    & $1.83 \pm 0.0014$
  & $6.81 \pm 0.0268$
 & $7.30 \pm 0.0256$
 &   $\bm{{0.38}} \pm 0.0017$
   \\
\rowcolor{Gray}
CTEG  & $262.19$     &$\bm{{9.96}}$ & $\bm{{1.22}} \pm 0.0020$ & $\bm{{8.18}} \pm 0.0508$ & $\bm{{9.35}} \pm 0.0430$ & $0.72 \pm 0.0055$ \\ 

\bottomrule
\end{tabular}%
}

\caption{Quantitative results on the ablation study. CTEG model achieves the best overall performance compared with other settings. More in-depth experiments are provided in the Appendix~\ref{appendix_cteg}.}
\label{CTEG_ablation}
\vspace{-2mm}
\end{table*}

\subsection{Experimental Settings}

We mainly utilize the EmoAva dataset to validate the CTEG. 
We also conduct experiments on existing listeners' dataset~\cite{lm_listener} for reference
in the Appendix~\ref{appendix_cteg}.
The settings of CTEG are detailed in \S \ref{CTEG_settings}.
For each comparing method, we randomly sample $50$ sentences from the test set and generate the corresponding expression sequence with a maximum length of $128$. 
As mentioned in \S \ref{consistency}, there are currently no suitable quantitative metrics for emotion-content consistency evaluation. 
Instead, we adopt perceptual experiments following~\citet{CVAE_GRU}.
Each sample is rendered as an avatar video at 24 frames per second and shown to five participants.
The participants are instructed to rank the outputs based on the emotional consistency between the facial expressions and the corresponding text.
The Fleiss’ kappa score for the average preference of the generated results is 0.77.


\vspace{-2mm}
\subsection{Baselines}
\paragraph{LM-Listener.} 
To the best of our knowledge, this is the only open-sourced method applicable to the text-to-3D expression task~\cite{lm_listener}.

We implement the model with their released code, with most parameters kept unchanged.
To ensure a fair comparison and obtain diverse outputs, we use \textit{top-p} sampling (\textit{top-p}=0.8) instead of greedy search.

\vspace{-2mm}
\paragraph{Shuffle.} Each expression sequence is randomly shuffled along the temporal axis to rigorously test the model’s sensitivity to temporal coherence and expression fluency.

\vspace{-2mm}
\paragraph{Random.} Following~\citet{lm_listener}, we also randomly select expression sequences from the training set to assess the model's ability to model the emotion-content consistency.

\vspace{-2mm}

\section{Experimental Results}
\subsection{Main results}

\begin{figure}[!t]
  \centering
  \begin{tikzpicture}[scale=0.8]
    \begin{axis}[
      ybar stacked,
      bar width=20pt,
      nodes near coords,
      every node near coord/.append style={font=\tiny}, 
      enlargelimits=0.09,
      enlarge y limits=0.01, 
      legend style={
        at={(0.5,1.0)},
        anchor=south,
        legend columns=-1,
        /tikz/every even column/.append style={column sep=5pt},
        draw=black,
        fill={rgb,255:red,248;green,248;blue,255},
        legend cell align=left,
        legend style={font=\scriptsize, draw=none, fill opacity=0.8, text opacity=1, rounded corners=2pt}
      },
       yticklabel style={font=\scriptsize},
      symbolic x coords={random,LM-Listener,w.o. LTA, w.o. Lg, w.o. EwA, CTEG,GT},
          xticklabels={
        {random},
        {LM-Listener},
        {w.o. LTA},
        {w.o. $\mathcal{L}_g$},
        {w.o. EwA},
        {CTEG},
        {GT}},
      xtick=data,
      x tick label style={font=\scriptsize,rotate=45,anchor=east,xshift=2pt,yshift=-5pt},
            cycle list={
        {fill={rgb,255:red,077; green,154; blue,199},draw={rgb,255:red,077; green,154; blue,199}},
        {fill={rgb,255:red,153; green,200; blue,224},draw={rgb,255:red,153; green,200; blue,224}},
        {fill={rgb,255:red,212; green,230; blue,239},draw={rgb,255:red,212; green,230; blue,239}},
        {fill={rgb,255:red,248; green,244; blue,242},draw={rgb,255:red,248; green,244; blue,242}},
        {fill={rgb,255:red,251; green,216; blue,195},draw={rgb,255:red,251; green,216; blue,195}},
        {fill={rgb,255:red,242; green,164; blue,129},draw={rgb,255:red,242; green,164; blue,129}},
        {fill={rgb,255:red,214; green,096; blue,077},draw={rgb,255:red,214; green,096; blue,077}}
      },
      every node near coord/.append style={/pgf/number format/fixed,
                        /pgf/number format/precision=3}
    ] 
    \addplot+[ybar] plot coordinates {(random,0.68) (LM-Listener,0.18) 
     (w.o. LTA,0.08) (w.o. Lg,0.06) (w.o. EwA,0.0) (CTEG,0.0) (GT,0.0)};
     \addlegendentry{Least preferred} 
    \addplot+[ybar] plot coordinates {(random,0.14) (LM-Listener,0.38) 
     (w.o. LTA,0.2) (w.o. Lg,0.16) (w.o. EwA,0.10) (CTEG,0.02) (GT,0.0)};
    \addlegendentry{} 
    \addplot+[ybar] plot coordinates {(random,0.02) (LM-Listener,0.08) 
     (w.o. LTA,0.22) (w.o. Lg,0.26) (w.o. EwA,0.16) (CTEG,0.18) (GT,0.08)};
    \addlegendentry{}
    \addplot+[ybar] plot coordinates {(random,0.0) (LM-Listener,0.02) 
     (w.o. LTA,0.26) (w.o. Lg,0.16) (w.o. EwA,0.2) (CTEG,0.36) (GT,0.0)};
    \addlegendentry{}
    \addplot+[ybar] plot coordinates {(random,0.1) (LM-Listener,0.0) 
     (w.o. LTA,0.12) (w.o. Lg,0.14) (w.o. EwA,0.38) (CTEG,0.24) (GT,0.02)};
    \addlegendentry{}
    \addplot+[ybar] plot coordinates {(random,0.06) (LM-Listener,0.34) 
     (w.o. LTA,0.10) (w.o. Lg,0.18) (w.o. EwA,0.16) (CTEG,0.16) (GT,0.0)};
    \addlegendentry{}
    \addplot+[ybar] plot coordinates {(random,0.0) (LM-Listener,0.0) 
     (w.o. LTA,0.02) (w.o. Lg,0.04) (w.o. EwA,0.0) (CTEG,0.04) (GT,0.9)};
    
    \addlegendentry{Most preferred}
    \end{axis}
\end{tikzpicture}

\vspace{-3mm}
\caption{A quantitative evaluation of user preferences regarding emotion-content consistency. The color bar from blue to red indicates preference levels from lowest to highest.
Expressions from CTEG better match text emotions than those from baselines.}
\label{CTEG_preference_study}
\end{figure}

As shown in Table \ref{CTEG_quantitative_all}, CTEG outperforms the baselines across all diversity metrics by a large margin.
We also visualize the diversity of expressions generated by CTEG in Figure \ref{CTEG_visualization_diversity} and \ref{appendix_diversity} (Appendix).
We randomly sample four sequences of expressions given a text.
These examples demonstrate that the generated expressions exhibit a rich diversity. 

Figure \ref{CTEG_preference_study} illustrates an evaluation of user preferences.
Participants are instructed to rank the expressions according to how well their emotions aligned with the input text.
Compared with the \textit{random} setting, LM-Listener, and CTEG, we find that the user preference for CTEG is higher than that of the other two methods.
This indicates the effectiveness of CTEG in modeling emotion-content consistency.

From Table \ref{CTEG_quantitative_all}, 
we observe that the \textit{Cppl} metric for the \textit{shuffle} setting is several orders of magnitude higher than that of the normal sequences, indicating CTEG's high sensitivity to the expression sequence order. 
The lower \textit{Cppl} value confirms CTEG's effectiveness in modeling expression smoothness.

 \begin{figure}[!t]
  \centering
  \definecolor{line1}{RGB}{221,133,84}
\definecolor{line2}{RGB}{76,114,176}
\definecolor{back}{RGB}{234,234,242}
\definecolor{box}{RGB}{216,216,219}

\begin{tikzpicture}[scale=0.95]
    \begin{axis}[
        ymax=1.4,
        ymin=0.0,
        xmax=57,
        xmin=0,
        ylabel={KL Term $\mathcal{L}_{KL}(t)$},
        y label style={font=\scriptsize, yshift=0cm},
        ytick={0,0.2,0.4,0.6,0.8,1.0,1.2,1.4,1.5},
        yticklabels={0,0.2,0.4,0.6,0.8,1.0,1.2,1.4},
        y tick label style = {font=\scriptsize, xshift=0cm,},
        xlabel={Training Steps},
        x label style={font=\small, yshift=0.1cm},
        xtick = {1,5,10,15,20,25,30,35,40,45,50,55},
        xticklabels={1,5,10,15,20,25,30,35,40,45,50,55},
        x tick label style = {font=\scriptsize, yshift=0.0cm},
        width=8.2cm,
        height=5.0cm,
        legend style={at={(0.835, 0.51)}, anchor=north, legend columns=1, font=\scriptsize},
        scaled ticks=false,
        xtick scale label code/.code={},
        xmajorgrids=true,
        ymajorgrids=true,
        axis line style={color=white}, 
        tick style={color=white}, 
        grid style={white,solid},
        legend style={
        fill=back, 
        draw=box,  
    },
    legend pos=north east,
        axis background/.style={fill=back},
    ]
        \addplot[thick, solid, line2, mark=*, mark size = 0.9pt, mark options={fill=line2}] coordinates{
(1, 0.784)
(3, 0.193)
(5, 0.137)
(7, 0.115)
(9, 0.099)
(11, 0.094)
(13, 0.095)
(15, 0.101)
(16, 0.099)
(21, 0.087)
(26, 0.081)
(31, 0.076)
(36, 0.071)
(41, 0.065)
(46, 0.059)
(51, 0.057)
(56, 0.053)
        };
        \addlegendentry{w.o. $\mathcal{L}_{g}$}
        
        \addplot[thick, dashed, line1, mark=x, mark size = 1.5pt, mark options={fill=line1, solid}] coordinates{
(1, 1.388)
(3, 0.579)
(5, 0.417)
(7, 0.344)
(9, 0.3)
(11, 0.263)
(13, 0.243)
(15, 0.232)
(16, 0.226)
(21, 0.188)
(26, 0.166)
(31, 0.152)
(36, 0.142)
(41, 0.133)
(46, 0.125)
(51, 0.122)
(56, 0.116)
        };
        \addlegendentry{w. $\mathcal{L}_{g}$}
    \end{axis}
\end{tikzpicture}
\vspace{-2mm}
  \caption{The effect of $\mathcal{L}_{g}$ loss (Eq.~\ref{eq:target_loss}) on the KL term in Eq.~\ref{loss_cvad}. $\mathcal{L}_{g}$ loss mitigates the rapid decrease of the KL term and prevents it from approaching zero.}
  \label{CTEG_kl}
\end{figure}

\subsection{Ablation Study}
We conduct a quantitative experiment on three key components (i.e., EwA module, LTA module and $\mathcal{L}_{g}$ loss function) in CTEG model. 
As shown in Table \ref{CTEG_ablation}, removing the EwA module results in a significant drop in the four diversity metrics (\textit{DoT}, \textit{FgD}, \textit{Diversity}, and \textit{MModality}).
This indicates that the EwA module makes a substantial contribution to the diversity of the generated sequences.

\begin{figure*}[!t]
  \centering
  \includegraphics[width=1.0\textwidth]{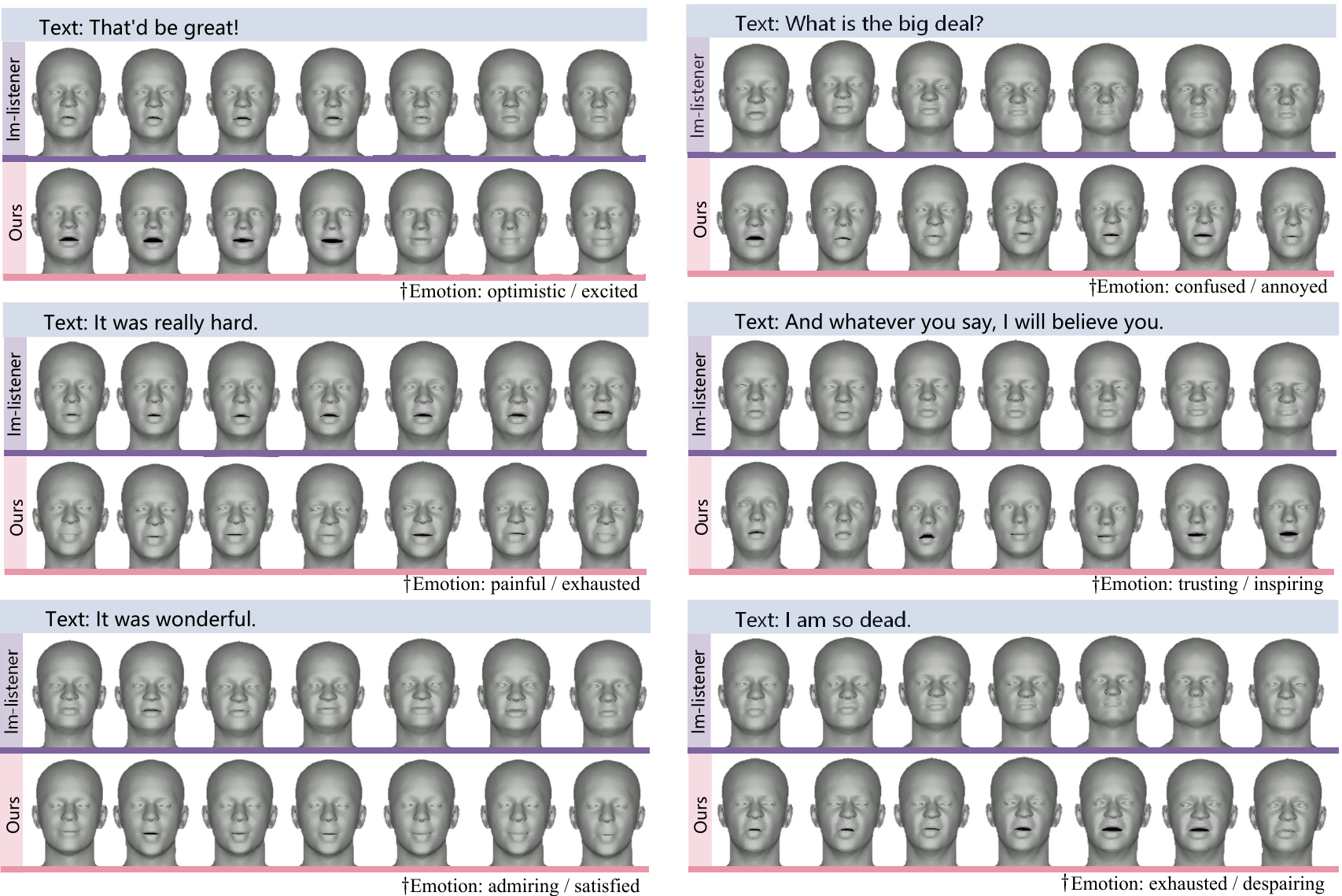}
  \caption{Qualitative analysis of the comparative results on emotion-content consistency. Our model demonstrates better consistency compared with the SoTA approach (LM-Listener). 
 $\dag$ represents the possible emotions conveyed by our results. More results are shown in the Figure~\ref{appendix_comparison}.
  }
  \label{CTEG_consistency}
\vspace{-3mm}
\end{figure*} 

From Table \ref{CTEG_ablation} and Figure \ref{CTEG_preference_study}, we can observe that removing the LTA module results in a decrease in emotion-content consistency compared with the full CTEG model. 
This highlights the importance of the LTA module and supports the assumptions of our method.
As shown in Table \ref{CTEG_ablation}, after removing the $\mathcal{L}{g}$ loss function, only the \textit{Variation} value shows improvement, while the performance of all other metrics declines. 
This indicates that removing the $\mathcal{L}{g}$ loss function leads to a drop in the overall performance of CTEG, and also
reflects a weakened fitting ability of CTEG. 
These experimental results indicate that the $\mathcal{L}_{g}$ loss function effectively mitigates this issue, enhancing the model's generalization ability and overall performance.

\vspace{-2mm}

\subsection{Discussion}
\vspace{-2mm}
\textbf{How does the $\mathcal{L}_{g}$ loss proposed in CTEG effectively mitigate the \textit{model collapse} problem?}
As shown in Figure \ref{CTEG_kl}, we plot the changes in the KL term in the loss function (Eq. \ref{CTEG_loss}) as the training steps progress. 
It can be observed that after removing $\mathcal{L}{g}$, the KL term quickly drops and approaches zero, whereas with $\mathcal{L}{g}$, the KL term decreases more gradually, and the curve remains consistently above that of the \textit{w.o. $\mathcal{L}_{g}$} setting as the training steps increase.
This phenomenon provides indirect support for the hypothesis proposed in \S \ref{target_loss}.

\begin{figure}[!t]
  \centering
  \begin{minipage}{0.49\columnwidth}
    \centering
    \includegraphics[width=\textwidth]{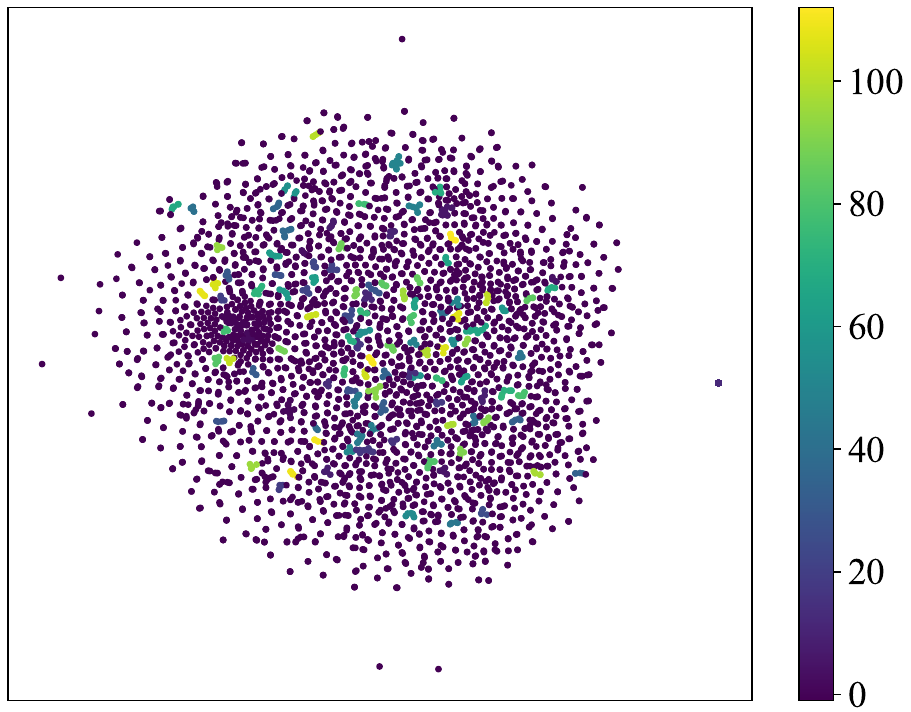}

  \end{minipage}
  \hfill
  \begin{minipage}{0.49\columnwidth}
    \centering
    \includegraphics[width=\textwidth]{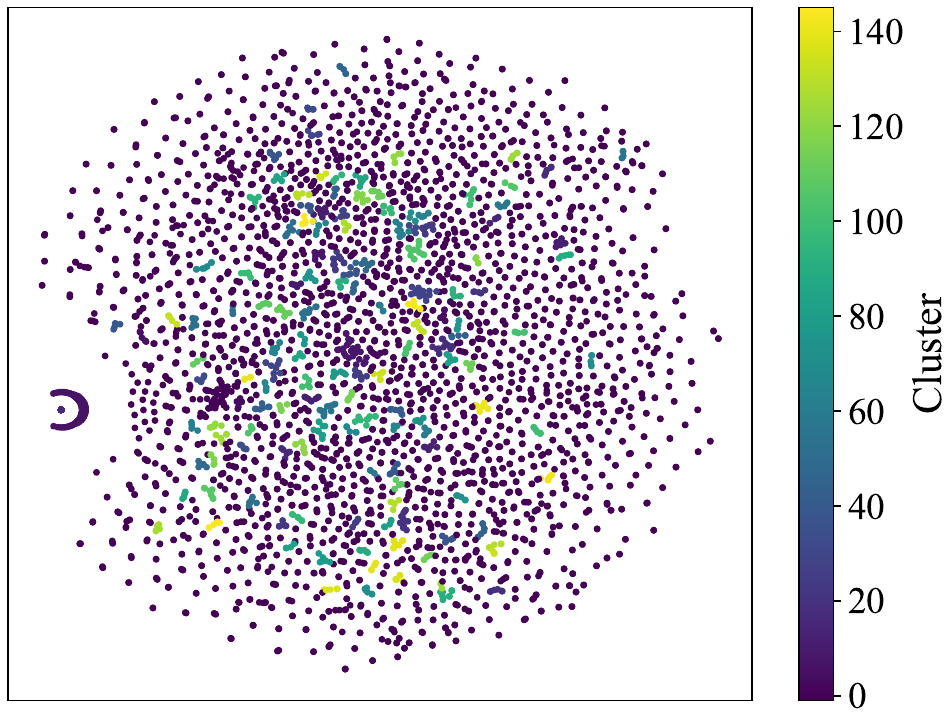}
  \end{minipage}
\caption{Comparison of latent variable distributions in the CVAD latent space. The \textit{w.EwA} model (right) captures a broader range of generative modes (146 clusters) than the \textit{w.o.EwA} model ( 113 clusters). This indicates that the EwA module enriches the feature space, enabling CVAD to model more diverse emotional patterns.}
  \label{CTEG_EwA}
  \vspace{-4mm}
\end{figure}

\textbf{Why does CTEG demonstrate stronger generative diversity in emotion?}
To understand why CTEG shows stronger diversity, we randomly sample from the latent space of CVAD and visualize the latent variable distribution (Figure \ref{CTEG_EwA}).
CTEG models more pattern clusters (146), a 29\% increase compared to the \textit{w.o. EwA} setting.
This confirms that the EwA module enriches input features, resulting in a more diverse latent space in CVAD. The latent variables capture more varied patterns, which improves the diversity of generated outputs.

\section{Conclusion} 
This paper proposes a novel end-to-end text-to-expression model, CTEG, which captures expressive variations in a continuous latent space, enabling the generation of diverse, fluid, and emotionally consistent facial expressions.
To support this task, we construct EmoAva, a large-scale and high-quality dataset consisting of 15,000 instances.
Extensive experiments demonstrate that CTEG significantly outperforms existing baselines across multiple aspects, paving the way for more emotionally aware digital humans.

\section*{Limitations}

\paragraph{Limited language coverage.}
Currently, the EmoAva dataset is limited to English due to resource constraints. Building a comprehensive multilingual text–expression dataset is inherently challenging, but we view it as a promising future direction. In the Appendix, we detail our data collection pipeline, which we hope will serve as a foundation for future expansion and community collaboration. Although human languages are diverse, emotional expression is largely universal. This motivates us to extend our work toward multilingual emotional expression modeling in future iterations.

\paragraph{Limited personalization or identity adaptation.}
CTEG is identity-agnostic by design, aiming to model generalizable human emotional states across speakers.
This design choice facilitates broader generalization but does not account for personalized expressive styles or speaker-specific traits. 
While personalized expression generation is an exciting future direction, we believe that building a strong foundation for modeling universal emotional patterns is a necessary first step—one that our work aims to establish.

\section*{Further Discussions}
\paragraph{The emotional uncertainty.}
Our work models facial expression diversity, which naturally involves a degree of uncertainty. 
Human emotions are inherently rich — the same sentence spoken by different individuals, or even by the same individual in different contexts, 
may come with entirely different expressions. 
Capturing this one-to-many mapping between language and expressions makes digital humans more engaging and lifelike.

\paragraph{Infinite expression diversity.} Our system can, in theory, generate an unlimited range of facial expressions, represented by continuous 53-dimensional expression coefficients obtained via PCA reduction in the FLAME framework.

\section*{Ethical Considerations}
\label{ethical}
\paragraph{Annotator compensation.} We employed three crowd-sourced annotators, all of whom are undergraduate students with strong English proficiency. They were compensated at an approximate rate of \$10 per hour, which aligns with standard local compensation for similar tasks.

\paragraph{Copyright and privacy.} We provide two licensing options for the dataset (detailed in ~\ref{appendix:license}), defining the conditions under which it may be accessed and used. The original video materials used to construct the dataset are sourced from publicly available television series, and their copyrights remain with the respective rights holders.
In addition, the facial features extracted for our model are identity-agnostic, meaning they do not retain personally identifiable characteristics of the actors. This serves as a form of de-identification, helping to preserve the portrait rights and privacy of the individuals appearing in the video content.

\section*{Acknowledgements}
We acknowledge supports from Guangdong Basic and Applied Basic Research Foundation 2025A1515012281, Nanjing Municipal Science and Technology Bureau 202401035 and University of Macau MYRG-GRG2024-00077-FST-UMDF.

\bibliography{custom}
\clearpage

\appendix
In this \textbf{appendix}, we present the following parts.
Section~\ref{appendix_dataset}: more details about the EmoAva dataset.
Section~\ref{appendix_related_work}: supplementary related work about dataset.
Section~\ref{appendix_cteg}: in-depth analysis about CTEG.
Section~\ref{appendix_metrics}: details and formulas of some evaluation metrics.
Section~\ref{visual_results}: additional visualization results.
Section~\ref{error_analysis}: analysis of failure cases.

\section{More Details about EmoAva Dataset}
\label{appendix_dataset}
Details of the dataset construction and license are provided here.
All processing code and the dataset will be made available via the provided anonymous GitHub repository.

\subsection{License}
\label{appendix:license}
We will provide two separate licenses for the dataset: one for the video files and another for the 3D expression code. The latter is directly relevant to the task presented in this paper and will be released under the CC BY-NC 4.0 license.
The video data, however, involves copyright considerations related to film and television content. Since such data can benefit a broader range of tasks beyond our primary focus, we plan to release it under a more restrictive license.
We have consulted legal experts regarding the conditions for releasing this type of data. The full licensing documentation will be made available in our GitHub repository.

\subsection{Construction Pipeline}
\begin{figure}[!t]
    \centering
    \includegraphics[width=\columnwidth]{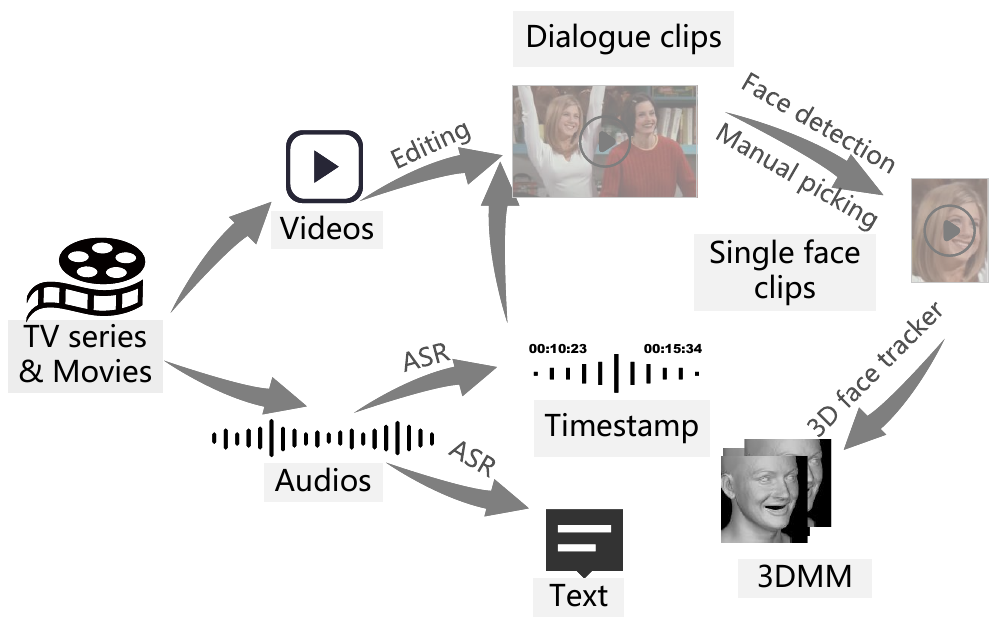}
    \caption{Pipeline for constructing the \textbf{EmoAva} dataset.}
    \label{fig:dataset_pipeline}
\end{figure}

The dataset construction pipeline is shown in Figure \ref{fig:dataset_pipeline}.
Given a raw video that may contain multiple people and varying shots, our goal is to extract a single talking-face video with a fixed camera view, and then extract 3D coefficients. 

To achieve the first step, we need to segment each raw video both spatially and temporally. 
Spatial segmentation involves isolating the talking face from multiple possibly co-occurring faces, while temporal segmentation involves extracting a continuous shot, typically where a person is speaking a complete sentence or segment.

Specifically, for a raw video, the audio is extracted and transcribed using speech recognition and ASR models (i.e., whisperX) \cite{whisperX}, to obtain timestamps and textual content. 
These timestamps typically correspond to complete utterances by individual actors. 
Utilizing these timestamps, we segment the raw video temporally, effectively achieving time-based segmentation.
Spatial segmentation, however, presents a more complex challenge. 

To our best knowledge, no current method can reliably identify the speaking face among multiple faces in a frame. 
For the videos lacking speaker identity information, we exclude frames containing multiple detected faces.
The construction algorithm involves multiple AI-based models, none of which can guarantee $100\%$ accuracy. 
To ensure the quality of the dataset, we perform a manual check on the segmented videos. 
This process achieves our initial objective.
After obtaining the segmented videos, we employ the widely-used FLAME tracking model EMOCA v2 \cite{emoca}, to extract 3D coefficients. Following this, we conduct a manual check on the final instances to ensure their accuracy and quality. 

In conclusion, we proposes a semi-automated approach that leverages several algorithms to generate large-scale instances. 
In this framework, human annotators are primarily tasked with verifying the algorithmic outputs and eliminating low-quality instances, thereby significantly enhancing efficiency and scalability.

\begin{table}[!t]
  
    \fontsize{8}{8.5}\selectfont
  \setlength{\tabcolsep}{5mm}

\begin{tabular}{lr} 
  \toprule
  \bf{{EmoAva}} & \bf Statistics\\
  \arrayrulecolor[gray]{0.5}\hline

  \# Train set & 12,000 \\
  \# Valid set & 1,500  \\
  \# Test set & 1,500 \\
 \# Dataset Total & 15,000 \\
  \hline
  Average sentence length & 14 \\
  Average expression steps & 52 \\
  Total frames of all expressions & 782,471 \\
  \hline
    \arrayrulecolor{black}
  \# One-to-many instances & 2270 \\
  \# Expressions exceeding 256 frames & 184 \\
  \bottomrule
  \end{tabular}
    \caption{Statistics of the EmoAva dataset.}
        \label{tab:statis}
\end{table}

\subsection{Guidelines for Human Annotation}

To maintain the data quality, we perform manual checking.
Specifically, we employ three annotators to remove low-quality instances, where the criteria are as follows: 
\textbf{1)} The face should be clearly visible, without obstructions like masks or sunglasses.
\textbf{2)} The actor's facial expression changes should be continuous (i.e., no scene cuts).
\textbf{3)} The actor should complete their sentence without being abruptly cut off.
\textbf{4)} There should be only one person in the video from start to finish.
\textbf{5)} The text should match what the actor is saying.
\textbf{6)} The avatar expressions (mesh format driven by tracked vectors) align with those in the corresponding videos.
We determine whether to drop data samples through independent annotation by the three annotators, followed by a majority vote on the results. 
After annotation, we calculate the Fleiss' kappa score~\cite{kappa}, achieving a value of $\mathbf{0.86}$. 
This indicates minimal disagreement among the annotators, reflecting the high quality of the dataset's annotations.


\subsection{Criteria for Collecting the Raw Videos}

Crucially, manual screening is necessary when selecting television show segments from the internet (i.e., YouTube). 
First, we need to avoid cartoons or fantasy genres that do not feature real human faces. 
Second, we must steer clear of videos that may contain violence, gore, or explicit content that is not appropriate for mainstream audiences.

\subsection{Dataset Insights}

We randomly partition all instances in the training set into three subsets: training/validation/test sets, comprising $80\%$/$10\%$/$10\%$ of the total, respectively. 
We provide a brief summary of the key characteristics of the EmoAva dataset in the following and in Table~\ref{tab:statis}.
\paragraph{Large-scale and High-quality.} To ensure data quality, we employ SoTA methods at every stage of the dataset preprocessing algorithm. 
Additionally, we manually check and remove the expressions that lack fluidity or do not consistently match the emotions expressed in the text.
As a result, our dataset comprises $15,000$ text-3D expression instances and a total of $782,471$ FLAME frames.
\paragraph{Diverse Mapping.} In a dataset of $15,000$ text-expression pairs, there are $2,270$ instances with a 1-to-N relationship (where N ranges from $2$ to $76$), accounting for over $15\%$.
\paragraph{Diverse Expressions.} EmoAva comes from a diverse range of sources and scenarios, i.e., including over $100$ movies and more than $5$ TV shows, thus resulting in a wide variety of expressions. 
\paragraph{Diverse Emotion.} Expressions exhibit a good diversity of emotions, including \textit{Happy}, \textit{Sad}, \textit{Neutral}, \textit{Disgust}, \textit{Fear}, \textit{Surprise}, and \textit{Angry}.\footnote{The statistical analysis is performed using a widely used emotion recognition framework DeepFace~\cite{deepface}.} 
\paragraph{Rich and Varied Emotions.} The emotions within a single sequence are also quite diverse, exhibiting significant variability.
The proportion of expressions containing two or more types of emotions exceeds $95\%$.
\paragraph{Highly Scalable.} The dataset includes raw videos, raw audio files, and a semi-automated construction algorithm, facilitating the extension to additional modalities and tasks.

\section{Supplementary Related Work}
\label{appendix_related_work}
\paragraph{3D Avatar Head Dataset.} 
This work also closely relates to the development and use of 3D Avatar Head Datasets. 
Numerous benchmark datasets~\cite{yin20063d,VOCA} exist within the community, but there is limited focus on benchmarks for emotion-aware dynamic 3D avatars.
Current dynamic 3D avatar benchmarks typically lack language signals (e.g., text or speech)~\cite{facedataset,ranjan2018generating,zhang2013high}, a gap our research aims to fill. 

To achieve Emotion-Content Consistency, an ideal 3D face dataset pairs each text instance with corresponding facial expression sequences as the text is articulated. 
Existing datasets~\cite{emotalk3D,zhang2016multimodal} often fall short, requiring annotators to read sentences with preset emotions, thereby ignoring the text's intrinsic emotional content. 
Notably, the MovieChat~\cite{face2face} dataset also includes 3D expression sequences.
Unfortunately, they only release FACS muscle features, which can not reconstruct realistic facial details and significantly diverge from the current research framework.

To overcome this, we collect 2D video clips of actors' dialogues from various films and TV scenes, closely mirroring real conversational scenarios with consistent emotion-language alignment. 
We then use the SoTA FLAME tracking approach to extract 3D expression codes and meshes from 2D videos.

\paragraph{Talking Head Video Datasets.}
As introduced before, our goal is to collect videos of talking faces that exhibit rich, emotionally varied expressions in naturalistic conversational settings. Several existing datasets are relevant, including those from the talking head synthesis~\cite{talkinghead1kh,VoxCeleb2,FaceForensics}, multimodal emotion analysis~\cite{MEMOR,MELD}, and text-video modeling~\cite{yu2023celebv}.

Among these, we find MELD~\cite{MELD} and MEMOR~\cite{MEMOR} to be the most suitable, as they are constructed from television show segments and contain real conversational dynamics with reasonably expressive faces. However, both datasets have notable limitations: they include only a small number of speakers, and their overall data scale is limited, which restricts their usefulness for training expressive generation models that aim for generalization and diversity.

In contrast, other commonly used datasets present further issues. For example, many talking head synthesis datasets involve individuals speaking directly to the camera with monotonous, emotionally flat expressions, lacking the nuanced variations seen in real social interaction. Additionally, several recent datasets collect videos from short-form content platforms (e.g., TikTok), but such self-recorded clips are often inconsistent in both expressive quality and visual fidelity, making them suboptimal for fine-grained expression modeling.

Based on this analysis, we find that movies and TV shows offer the most suitable source material, as they combine diverse emotional content, high production quality, and natural dialogue. 
To address the limitations in scale and speaker diversity of MELD and MEMOR, we construct a new dataset, further augmenting it with high-quality conversational clips sourced from YouTube. 
This hybrid strategy ensures a more scalable and emotionally rich dataset for expressive facial behaviors.

\section{In-depth Analysis of CTEG}
\label{appendix_cteg}
In this section, we present more experimental results about the analysis of CTEG.
Specifically, we investigate the following aspects:
\begin{tcolorbox}[breakable, mybox]
{\small
\begin{enumerate}[label=\ding{72},leftmargin=0pt, itemsep=0.5ex]
  \item \textbf{Q1}: How does CTEG perform when evaluated on the LM-Listener dataset?
  \item \textbf{Q2}: Should the projection layer be shared between the input and the output of the CVAD module?
  \item \textbf{Q3}: Can the attention mechanism in the LTA module be replaced with a simpler average pooling operation?
    \item \textbf{Q4}: Should the weights of the pretrained language model (i.e., BERT) be fixed?
  \item \textbf{Q5}: Does increasing the number of CVAD layers yield better performance?
\end{enumerate}
}
\end{tcolorbox}

\begin{table}[!t]

\centering
\footnotesize
\resizebox{1.0\linewidth}{!}{%
\small
\begin{tabular}{lllll}
\toprule
            & L2 $\downarrow$        & FD $\downarrow$         & Variation $\rightarrow$     & P-FD $\downarrow$       \\ \cmidrule{2-5}
\rowcolor{light-gray}
\textit{GT}       & /             & /             & $0.11$           & /              \\
LM-Listener       & $0.43 \pm 0.02$ & $18.22 \pm 0.70$ & $0.116 \pm 0.005$ & $19.63 \pm 0.80$ \\
\rowcolor{Gray}
CTEG              & $\bm{0.37} \pm 0.03$ & $\bm{16.92} \pm 1.20$ & $\bm{0.114} \pm 0.007$  & $\bm{16.55} \pm 0.90$  \\
\bottomrule
\end{tabular}%
}
\caption{Quantitative results on the LM-Listener dataset.}
\label{tab:LM-Listener}
\end{table}

\subsection{More Details and Experimental Settings}
We make several variants of CTEG in this paper.
\textit{w. sharing} refers to the model that shares the projection layers between the input and output of the CVAD module.
Compared with it, CTEG does not share the projection layers.
Based on \textit{w. sharing}, we also test the performance of two variants, \textit{w. pooling} and \textit{w. BERT fine-tuning}. 
Compared to \textit{w. sharing}, \textit{w. pooling} modifies the LTA module by replacing the attention operation with an average pooling operation (i.e., averaging the latent variables from time steps \( 1 \) to \( s \) for time step \( s \)).
Compared to \textit{w. sharing}, \textit{w. BERT fine-tuning} involves fine-tuning the BERT model during the training process, while the former fixes the weights of BERT.
In addition to these, we also provide the performance of the ground truth (GT) on certain metrics for reference.

Our system is lightweight, with a total parameter count under 130M. 
All experiments in this paper are conducted on a single NVIDIA A100 GPU.

\subsection{Results and Analysis}

\paratitle{\(\blacktriangleright\) Q1: How does CTEG perform when evaluated on the LM-Listener dataset?}

We conduct a comparative experiment on the dataset used in~\citet{lm_listener}, following exactly the same evaluation settings. The results are presented in Table~\ref{tab:LM-Listener}. As shown, CTEG consistently outperforms the LM-Listener model across all metrics, demonstrating its strong regression capability in the text-to-expression generation task.

Notably, we do not apply our proposed evaluation metrics to their dataset due to a fundamental mismatch in data characteristics. Specifically, their dataset lacks one-to-many mappings and includes only a single character. As a result, it does not account for emotional diversity—each input text corresponds to only one facial expression.

\begin{table*}[!t]

\centering
\scriptsize
\resizebox{0.98\linewidth}{!}{%
\tiny
\begin{tabular}{lllllll}
\toprule
    & Cppl $\downarrow$      & DoT $\rightarrow$      & FgD $\rightarrow$      & Diversity $\uparrow$  & MModality $\uparrow$  & Variation $\rightarrow$  \\ \cmidrule{2-7}
\rowcolor{light-gray}
\textit{GT} & / & $9.24$     &$1.26$     & /     & /     & $0.28$     \\
w. sharing    & $\bm{241.28}$         & $\bm{{8.53}}$ & $\bm{{1.27}} \pm 0.0032$ & $7.16 \pm 0.0436$ &  ${{8.01}} \pm 0.0292$ & $0.53 \pm 0.0027$ \\ 
\quad w. pooling   & $300.10$             & $7.89$          & $1.30 \pm 0.0102$          & $3.17 \pm 0.0225$          & $6.47 \pm 0.0605$          &  $\bm{0.32} \pm 0.0052$          \\
\quad w. BERT fine-tuning & $249.14$ & $6.28$          & $1.10 \pm 0.0022$          & $6.29 \pm 0.0552$          & $6.28 \pm 0.0399$          & $0.60 \pm 0.0063$          \\

\rowcolor{Gray}
CTEG  & $262.19$     &${{9.96}}$ & ${{1.22}} \pm 0.0020$ & $\bm{{8.18}} \pm 0.0508$ & $\bm{{9.35}} \pm 0.0430$ & $0.72 \pm 0.0055$ \\ 

\bottomrule
\end{tabular}%
}
\caption{Quantitative results on some variants of CTEG. 
Lower values ($\downarrow$) and higher values ($\uparrow$) are preferred, while values closer to the Ground Truth (\textit{GT}) are indicated by $\rightarrow$.
The standard error is estimated through bootstrap resampling with $1000$ iterations.}

\label{CTEG_variant}
\end{table*}

\paratitle{\(\blacktriangleright\) Q2: Should the projection layer be shared between the input and the output of the CVAD module?}

As shown in Table \ref{CTEG_variant}, \textit{w. sharing} achieves the best results on the Cppl, DoT, and FgD metrics. 
The \textit{Variation} also outperforms \textit{CTEG} (w.o.sharing), but the \textit{Diversity} and \textit{MModality} metrics are lower than those of \textit{CTEG}. 
Upon closer inspection, we find that the improvements in the Cppl and DoT metrics under the \textit{w. sharing} setting are minimal. 
The difference between \textit{CTEG} and \textit{GT} for the \textit{DoT} metric is $0.72 (9.96 - 9.24)$, while the difference between \textit{w. sharing} and \textit{GT} is $0.71 (9.24 - 8.53)$. 
One possible explanation for this phenomenon is the constraint on the generated vector space introduced by sharing the input and output mapping layers, which reduces diversity and yields metrics similar to those of \textit{GT}.

Although \textit{w. sharing} closely approximates GT on some metrics, we still remove the sharing operation in CTEG. 
The reason is that GT metrics serve only as reference values, and there are inherent differences between the test set used to compute GT and real-world data. Therefore, we can not use GT metrics (i.e., DoT, FgD, and Variation) as the sole criterion for evaluating the quality of our methods. 
In contrast, the performance of \textit{w. sharing} on Diversity and MModality metrics is significantly inferior to that of CTEG. 
Considering these trade-offs, we believe that the sharing operation should not be retained in CTEG.

\usetikzlibrary{intersections}  
\usepgfplotslibrary{fillbetween} 
\definecolor{back2}{RGB}{234,234,242}

\paratitle{\(\blacktriangleright\) Q3: Can the attention mechanism in the LTA module be replaced with a simpler average pooling operation?}

When comparing \textit{w. pooling} and \textit{w. sharing} in Table \ref{CTEG_variant}, we observe a downward trend in many diversity metrics. This indicates that the attention operation in the LTA module is more effective than the pooling operation, which aligns with our intuition. 
For each current time step, a more reasonable integration of historical attention is beneficial, while simply averaging historical states may diminish the meaningfulness of the feature representation at the current moment, thereby weakening its rich representational capacity.

\paratitle{\(\blacktriangleright\) Q4: Should the weights of the pretrained language model (i.e., BERT) be fixed?}

In the comparison between \textit{w. BERT fine-tuning} and \textit{w. sharing}, almost all metrics showed a decline in performance in Table \ref{CTEG_variant}. 
We think this may be due to the limited size of the training set, which could have led to overfitting when fine-tuning the text embedding model.

\begin{figure}[!t]
    \centering

    \begin{tikzpicture}
    \begin{axis}[
        hide axis,
        xmin=0, xmax=1,
        ymin=0, ymax=1,
        legend columns=2,
        legend style={at={(0.0,-0.1)}, anchor=north, draw=none, font=\scriptsize,column sep=5pt},
        ]
        \addplot[thick, solid, color=c1] coordinates {(0,0)};
        \addlegendentry{Performance}
        
        \addplot[thick, dashed, color=c5] coordinates {(0,0)};
        \addlegendentry{Ground Truth}
    \end{axis}
    \end{tikzpicture}
    
    \resizebox{0.9\columnwidth}{!}{
    \begin{minipage}{0.35\columnwidth}
    \begin{tikzpicture}
    \begin{axis}[
        ymax=500,
        ymin=200,
        xmax=5.5,
        xmin=0.6,
        ylabel={CPerplexity},
        y label style={font=\scriptsize, yshift=-0.2cm},
        ytick={200,250, 300,350, 400,450, 500},
        yticklabels={200, 250,300, 350,400,450, 500},
        y tick label style = {font=\scriptsize, xshift=0.1cm,},
        xlabel={Decoder layer},
        x label style={font=\scriptsize, yshift=0.1cm},
        xtick = {0,1,2,3,4,5},
        xticklabels={0, 1,2,3,4,5},
        x tick label style = {font=\scriptsize, yshift=0.0cm},
        width=4.0cm,
        height=4.0cm,
        legend style={at={(0.835, 0.51)}, anchor=north, legend columns=1, font=\scriptsize},
        xmajorgrids=true,
        ymajorgrids=true,
        grid style={white,solid},
        tick style={color=white},
        axis line style={color=white},
        axis background/.style={fill=back2},
    ]
    
        \addplot[thick, solid, c1, mark=o, mark size = 1.2pt, mark options={fill=c1, solid}] coordinates{
            (1, 241.28) (2, 294.12) (3, 289.51) (4, 487.51)
        };
        
    \end{axis}
    \end{tikzpicture}
    \end{minipage}
    \hfill
    \begin{minipage}{0.33\columnwidth}
    \begin{tikzpicture}
    \begin{axis}[
        ymax=10,
        ymin=0,
        xmax=5.5,
        xmin=0.6,
        ylabel={MModality},
        y label style={font=\scriptsize, yshift=-0.3cm},
        ytick={0, 2, 4, 6,8,10},
        yticklabels={0,2,4,6,8,10},
        y tick label style = {font=\scriptsize, xshift=0.1cm,},
        xlabel={Decoder layer},
        x label style={font=\scriptsize, yshift=0.1cm},
        xtick = {0,1,2,3,4,5},
        xticklabels={0, 1,2,3,4,5},
        x tick label style = {font=\scriptsize, yshift=0.0cm},
        width=4.0cm,
        height=4.0cm,
        legend style={at={(0.835, 0.51)}, anchor=north, legend columns=1, font=\scriptsize},
        xmajorgrids=true,
        ymajorgrids=true,
        grid style={white,solid},
        tick style={color=white},
        axis line style={color=white},
        axis background/.style={fill=back2},
    ]
    
        \addplot[thick, solid, c1, mark=o, mark size = 1.2pt, mark options={fill=c1, solid}] coordinates{
            (1, 8.01) (2, 6.11) (3, 8.59) (4, 5.40) (5,1.05) 
        };
        
    \end{axis}
    \end{tikzpicture}
    \end{minipage}
    \hfill
    \begin{minipage}{0.33\columnwidth}
        \begin{tikzpicture}
    \begin{axis}[
        ymax=8,
        ymin=0,
        xmax=5.5,
        xmin=0.6,
        ylabel={Diversity},
        y label style={font=\scriptsize, yshift=0.0cm},
        ytick={0, 1,2,3, 4,5, 6,7,8},
        yticklabels={0,1,2,3,4,5,6,7,8},
        y tick label style = {font=\scriptsize, xshift=0.1cm,},
        xlabel={Decoder layer},
        x label style={font=\scriptsize, yshift=0.1cm},
        xtick = {0,1,2,3,4,5},
        xticklabels={0, 1,2,3,4,5},
        x tick label style = {font=\scriptsize, yshift=0.0cm},
        width=4.0cm,
        height=4.0cm,
        legend style={at={(0.835, 0.51)}, anchor=north, legend columns=1, font=\scriptsize},
        xmajorgrids=true,
        ymajorgrids=true,
        grid style={white,solid},
        tick style={color=white},
        axis line style={color=white},
        axis background/.style={fill=back2},
    ]
    
        \addplot[thick, solid, c1, mark=o, mark size = 1.2pt, mark options={fill=c1, solid}] coordinates{
            (1, 7.16) (2, 4.11) (3, 5.32) (4, 4.42) (5,1.04) 
        };
        
    \end{axis}
    \end{tikzpicture}
    \end{minipage}
    }
    \vspace{1cm} 
    \resizebox{0.9\columnwidth}{!}{
    \begin{minipage}{0.35\columnwidth}
    \begin{tikzpicture}
    \begin{axis}[
        ymax=1.3,
        ymin=1.0,
        xmax=5.5,
        xmin=0.6,
        ylabel={FgD},
        y label style={font=\scriptsize, yshift=-0.25cm},
        ytick={1.0,1.05, 1.1, 1.15, 1.2,1.25, 1.3},
        yticklabels={1.0,1.05, 1.1, 1.15, 1.2,1.25, 1.3},
        y tick label style = {font=\scriptsize, xshift=0.1cm,},
        xlabel={Decoder layer},
        x label style={font=\scriptsize, yshift=0.1cm},
        xtick = {0,1,2,3,4,5},
        xticklabels={0, 1,2,3,4,5},
        x tick label style = {font=\scriptsize, yshift=0.0cm},
        width=4cm,
        height=4cm,
        legend style={at={(0.835, 0.51)}, anchor=north, legend columns=1, font=\scriptsize},
        xmajorgrids=true,
        ymajorgrids=true,
        grid style={white,solid},
        tick style={color=white},
        axis line style={color=white},
        axis background/.style={fill=back2},
    ]
    
        \addplot[thick, solid, c1, mark=o, mark size = 1.2pt, mark options={fill=c1, solid}] coordinates{
            (1, 1.267) (2, 1.241) (3, 1.171) (4, 1.281) (5,1.048) 
        };
        \addplot[thick, dashed, color=c5] coordinates {(0, 1.255) (6, 1.255)};
        
    \end{axis}
    \end{tikzpicture}
    \end{minipage}
    \hfill
    \begin{minipage}{0.333\columnwidth}
    \begin{tikzpicture}
    \begin{axis}[
        ymax=10,
        ymin=2,
        xmax=5.5,
        xmin=0.6,
        ylabel={DoT},
        y label style={font=\scriptsize, yshift=-0.2cm},
        ytick={2,3,4,5,6,7,8,9,10},
        yticklabels={2,3,4,5,6,7,8,9,10},
        y tick label style = {font=\scriptsize, xshift=0.1cm,},
        xlabel={Decoder layer},
        x label style={font=\scriptsize, yshift=0.1cm},
        xtick = {0,1,2,3,4,5},
        xticklabels={0, 1,2,3,4,5},
        x tick label style = {font=\scriptsize, yshift=0.0cm},
        width=4.0cm,
        height=4.0cm,
        legend style={at={(0.835, 0.51)}, anchor=north, legend columns=1, font=\scriptsize},
        xmajorgrids=true,
        ymajorgrids=true,
        grid style={white,solid},
        tick style={color=white},
        axis line style={color=white},
        axis background/.style={fill=back2},
    ]
    
        \addplot[thick, solid, c1, mark=o, mark size = 1.2pt, mark options={fill=c1, solid}] coordinates{
            (1, 8.525) (2, 6.551) (3, 9.000) (4, 6.479) (5,2.139) 
        };
        \addplot[thick, dashed, color=c5] coordinates {(0, 9.241) (6, 9.241)};
        
    \end{axis}
    \end{tikzpicture}
    \end{minipage}
    \hfill
    \begin{minipage}{0.33\columnwidth}
        \begin{tikzpicture}
    \begin{axis}[
        ymax=0.6,
        ymin=0.0,
        xmax=6,
        xmin=0,
        ylabel={Variation},
        y label style={font=\scriptsize, yshift=-0.15cm},
        ytick={0.0,0.1,0.2,0.3,0.4,0.5,0.6},
        yticklabels={0.0,0.1,0.2,0.3,0.4,0.5,0.6},
        y tick label style = {font=\scriptsize, xshift=0.1cm,},
        xlabel={Decoder layer},
        x label style={font=\scriptsize, yshift=0.1cm},
        xtick = {0,1,2,3,4,5},
        xticklabels={0, 1,2,3,4,5},
        x tick label style = {font=\scriptsize, yshift=0.0cm},
        width=4.0cm,
        height=4.0cm,
        legend style={at={(0.835, 0.51)}, anchor=north, legend columns=1, font=\scriptsize},
        xmajorgrids=true,
        ymajorgrids=true,
        grid style={white,solid},
        tick style={color=white},
        axis line style={color=white},
        axis background/.style={fill=back2},
    ]
    
        \addplot[thick, solid, c1, mark=o, mark size = 1.2pt, mark options={fill=c1, solid}] coordinates{
            (1, 0.53) (2, 0.15) (3, 0.11) (4, 0.10) (5,0.01) 
        };
        \addplot[thick, dashed, color=c5] coordinates {(0, 0.28) (6, 0.28)};
        
    \end{axis}
    \end{tikzpicture}
    \end{minipage}
    }
    \vspace{-5mm}
    \caption{The performance of all three metrics shows a downward trend with the increasing number of decoder layers.}

    \label{decoder-layer}
    \end{figure}

\paratitle{\(\blacktriangleright\) Q5: Does increasing the number of CVAD layers yield better performance?}

We set the number of decoder layers from $1$ to $5$
to observe the diversity and naturalness of the generated
expressions.  
As shown in Figure \ref{decoder-layer}, it is evident that as the number of decoder layers increases, the model’s performance gradually declines.
Particularly when the number of layers reaches $5$, the perplexity explodes, increasing by several orders of magnitude compared to the 4-layer decoder, and the diversity also
becomes very poor. 
We find that the model struggles to
converge under many-layer conditions. We speculate that this is because each step $s$ in every layer is independently sampled, and $N$ layers would generate $s^N$ latent variable
states, introducing too much randomness. 
We refer to this phenomenon as Cumulative Sampling Instability. Therefore, for the method described in this paper, using a single-layer
decoder is the optimal configuration.

\begin{figure*}[!t]
  \centering
  \includegraphics[width=1.0\textwidth]{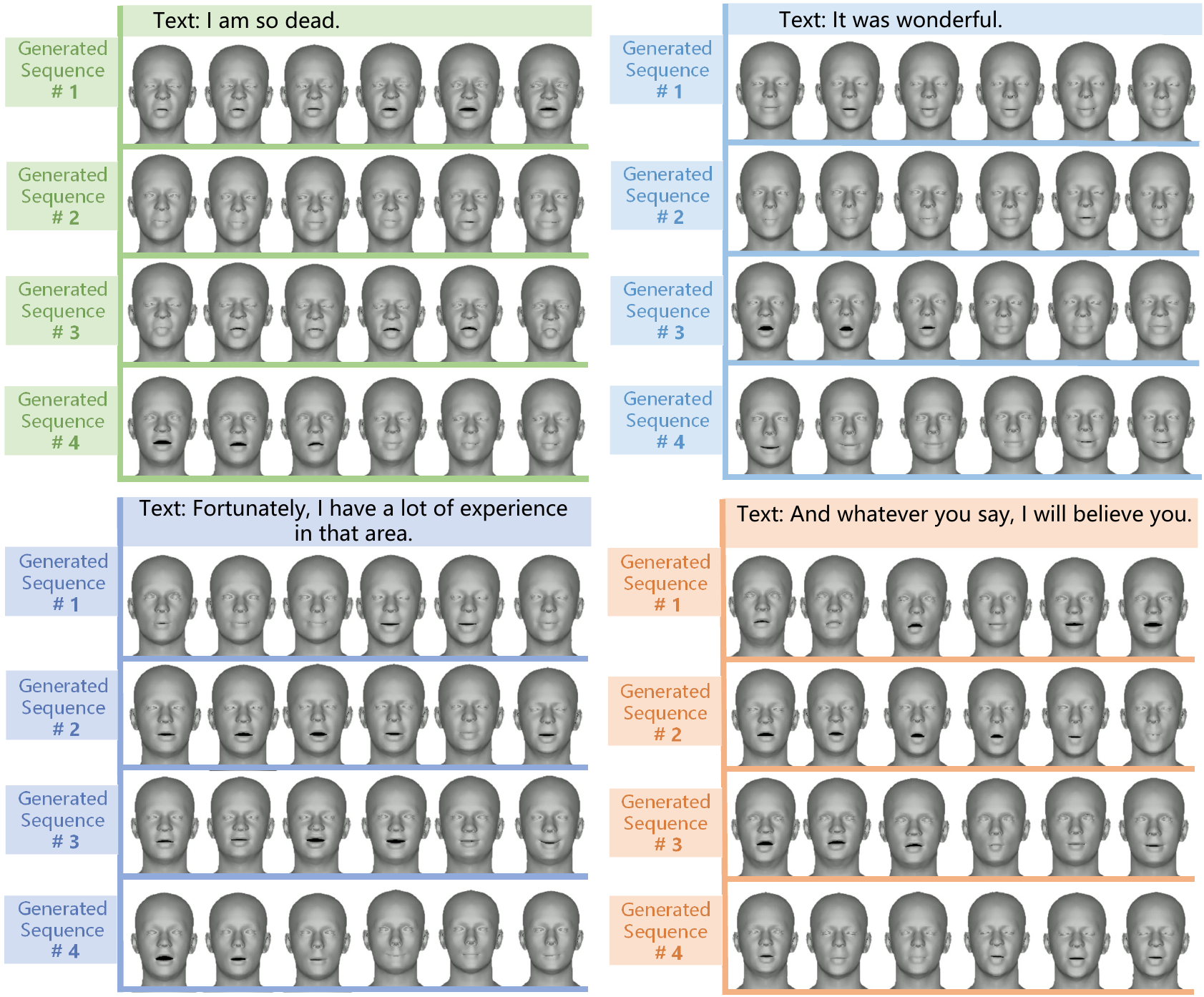}
  \caption{Visualization of the diversity generated by the CTEG model. Four sequences of expressions are generated from the same text with different random seeds. CTEG exhibits excellent generative diversity.}
  \label{CTEG_visualization_diversity}
\vspace{-2mm}
\end{figure*}



\section{Details of Evaluation Metrics}
\label{appendix_metrics}

\paragraph{Diversity} metric is calculated by the following formula \cite{T2M-GPT}:
\begin{equation}
\mathrm{Diversity}  =  \frac{1}{N_{d}}  \sum_{i=1}^{N_{d}} \left\| \Psi_{i}  - \Psi_{i}^{'} \right\| \,,
\end{equation}
where $\Psi$ and $\Psi'$ denote a pair of randomly sampled sequences of expression vectors that are generated without giving any text.
We set $N_{d}$ to $750$ in this paper. 
\paragraph{MModality} is calculated by the following formula \cite{T2M-GPT}:
\begin{equation}
\mathrm{MModality}  = \frac{1}{N_{m}} \sum_{i=1}^{N_{m}} \left\| \psi_{i}       - \psi_{i}^{'} \right\| \,.
\end{equation}
We set $N_{m}$ to $1500$ in this paper.
$\psi_{i}$ and $\psi'_{i}$ represent two different sequences generated under the same set of given texts.
\paragraph{Variation} is calculated by the following formula \cite{lm_listener}:
\begin{equation}
\mathrm{Variation} = \frac{1}{N_v} \sum_{i=1}^{N_v} \left( \frac{1}{n_i} \sum_{j=1}^{n_i} \mathrm{var}(\mathrm{E}_{ij}) \right) \,,
\end{equation}
where $\mathrm{E}_{ij}$ denotes a frame in a sequence of expression vectors.
$n_{i}$ is the length of the $i$-th sequence.
$N_{v}$ here is the number of sequences, which is set to $1500$.
$\mathrm{var(\cdot)}$ operation calculates the variance.

\paragraph{Fine-grained Diversity (FgD)} is calculated by:
\begin{equation}
  \text{FgD} = \frac{1}{(T-1)N} \sum_{i=1}^{N}\sum_{j=0}^{T-1} \left\| \mathbf{E}_{i,j+1} - \mathbf{E}_{i,j} \right\| \,.
\end{equation}

\paragraph{Diversity on Test (DoT)} is obtained by calculating the average Euclidean distance between each pair of generated expression sequences:
\begin{equation}
  \text{DoT} = \frac{2}{N(N-1)} \sum_{1 \leq i < j \leq N} \| \mathbf{E}_i - \mathbf{E}_j \| \,.
\end{equation}

\paragraph{Continuous perplexity (Cppl).}
Given the $i$-th expression sequence $\psi^{i}$, we define the following entropy inspired by \citet{perplexity}.
\begin{equation}
  H_{i}(\xi) \approx -\frac{1}{T}\sum_{j=1}^{T} \log_{2} p_{\xi}(\psi_{j}^i \mid \psi_{<j}^i, \mathbf{x}) \,,
\end{equation}
$p_{\xi}$ here is a continuous conditional distribution where modeled by a generation model.
Multivariate normal distribution is adopted in this paper and it is calculated by:
\begin{align}
  p_{\xi }(\psi_{j} \mid \psi_{<j}, \mathbf{x}) 
  &\approx \Phi (x+\delta;\mathbf{\mu}_{\xi}^j,\sigma^2\mathbf{I}) \notag \\
  &\quad - \Phi (x-\delta;\mathbf{\mu}_{\xi}^j,\sigma^2\mathbf{I}) \,,
\end{align}

where $\Phi(\cdot;\mathbf{\mu_{\xi}},\sigma^2\mathbf{I})$ denotes the cumulative distribution function (CDF) of the multivariate normal distribution with mean $\mu$ and covariance matrix $\sigma^2\mathbf{I}$.
Note that $\delta$ and $\sigma$ are empirical values, which are set to $0.8$ and $0.2$ here.

Given $N$ expression sequences, Cppl is calculated by:
\setlength{\abovedisplayskip}{0pt}
\setlength{\belowdisplayskip}{0pt}
\setlength{\abovedisplayshortskip}{0pt}
\setlength{\belowdisplayshortskip}{0pt}
\begin{equation}
  Cppl = 2^{\frac{1}{N}\sum_{i=1}^{N} H_{i}(\xi) } \,.
\end{equation}

\section{More Visualization Results}
\label{visual_results}
\begin{figure*}[!t]
  \centering
  \includegraphics[width=1.0\textwidth]{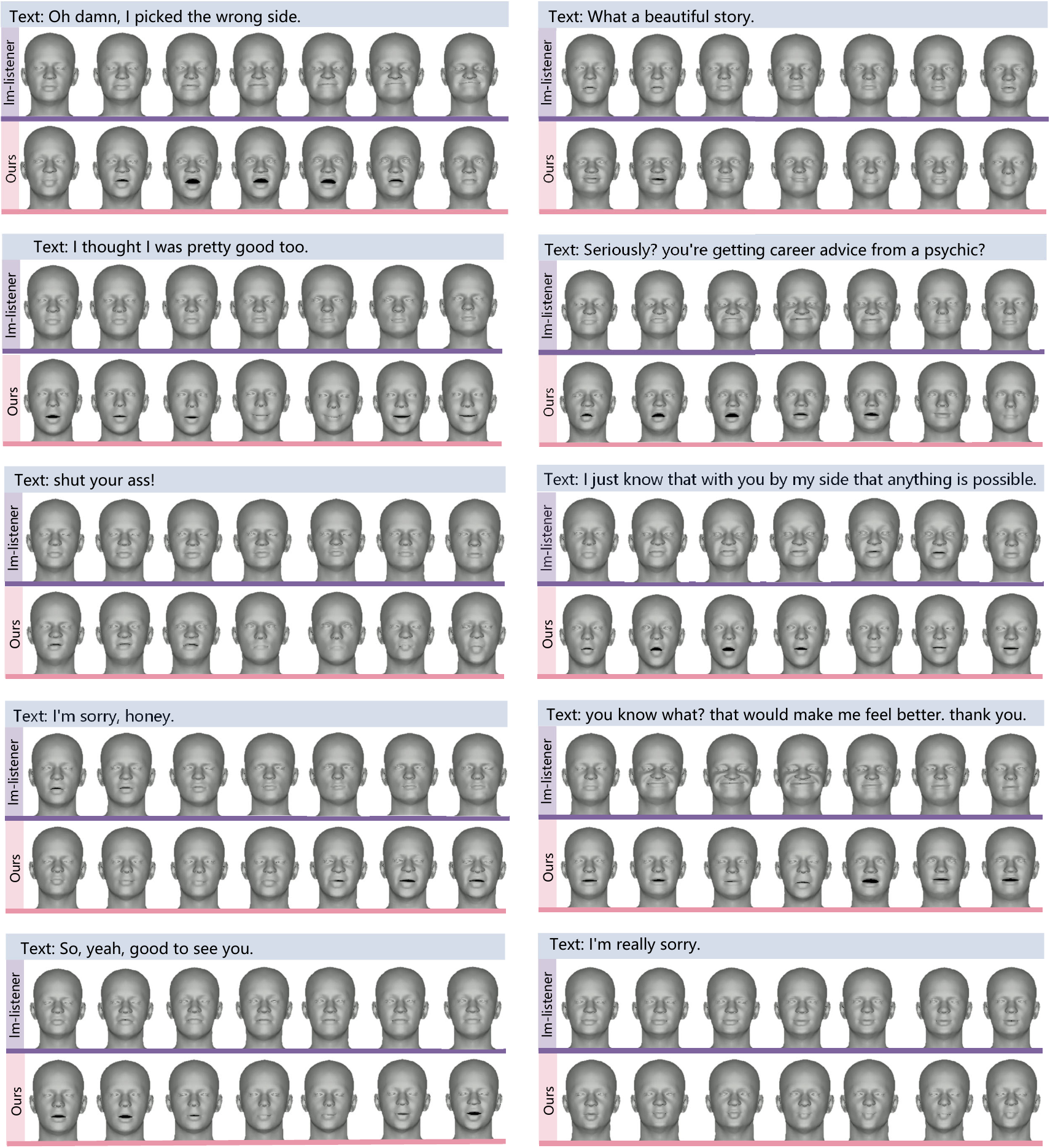}
\caption{Extended visual results generated by CTEG and LM-Listener \cite{lm_listener}. The expressions produced by CTEG exhibit greater consistency with the emotions conveyed by the corresponding text.}
  \label{appendix_comparison}
\end{figure*}

In this section we present extended visualization results of the expressions generated by CTEG and the baseline method \cite{lm_listener}.
As shown in Figure \ref{appendix_comparison}, we present several sequences of expressions generated by CTEG and LM-Listener \cite{lm_listener}.
Compared with the LM-Listener, the expressions we generated exhibit a greater alignment with the emotions conveyed by the corresponding text.
For example, the text ``Oh damn, I picked the wrong side.'' conveys emotions of pain, regret, and complaint. 
The expressions we generated effectively reflect these emotions. 
In contrast, the expressions produced by LM-Listener appear to convey a smile, which is inconsistent with the emotional tone of the text.
A similar observation is evident in the text ``What a beautiful story.'' 
This statement conveys feelings of joy and admiration, and the expressions we generated reflect this joy. 
However, the expressions produced by LM-Listener appear to convey a sense of indifference.
Many additional cases also support this observation in Figure \ref{appendix_comparison}.

\begin{figure*}[!t]
  \centering
  \includegraphics[width=1.0\textwidth]{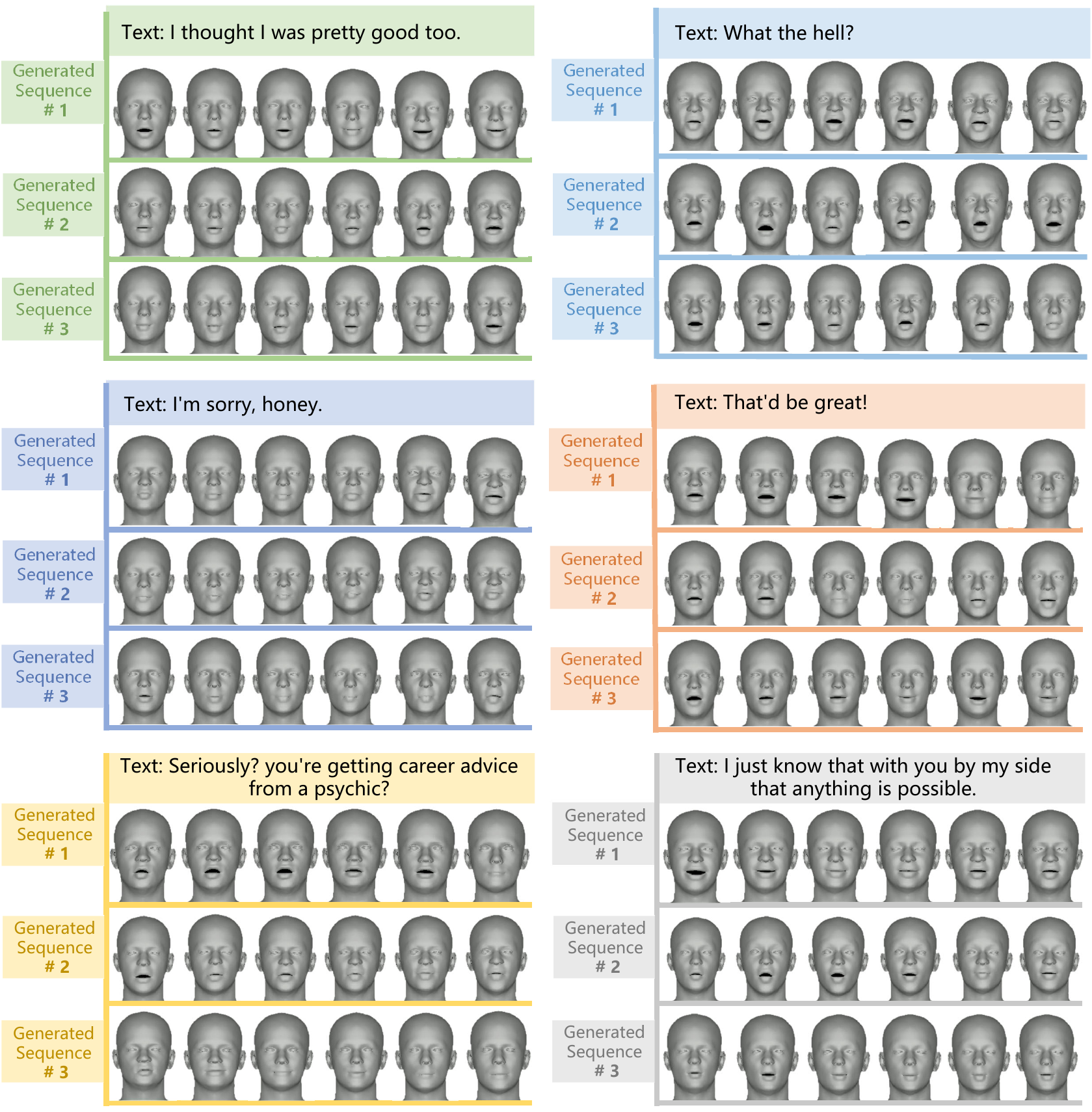}
  \caption{Visualization of the diversity generated by the CTEG model. Three sequences of expressions are generated from the same text with different random seeds. CTEG exhibits excellent generative diversity.}
  \label{appendix_diversity}
\end{figure*}

Figure~\ref{CTEG_visualization_diversity} and \ref{appendix_diversity} present the visual expressions generated by CTEG with several different random seeds.
Taking the text ``I thought I was pretty good too.'' as an example, this statement conveys emotions of pride, joy, or happiness. 
In the first sequence of expressions, the portrayal is one of delight. 
The second sequence exhibits a more subdued happiness, while the third sequence conveys a sense of pride and self-satisfaction. 
Despite the significant differences among these three sequences of expressions, all align well with the emotions conveyed by the text.
Similarly, taking the text ``What the hell?'' as an example, this statement generally conveys feelings of surprise, frustration, or disbelief. 
The three generated sequences of expressions all seem to convey these emotions. 
However, each sequence exhibits varying degrees of intensity in expressing surprise or frustration.
In addition to these, this phenomenon can also be observed in other examples.

\section{Error Analysis}
\label{error_analysis}

\begin{figure}[!t]
  \centering
  \includegraphics[width=0.8\columnwidth]{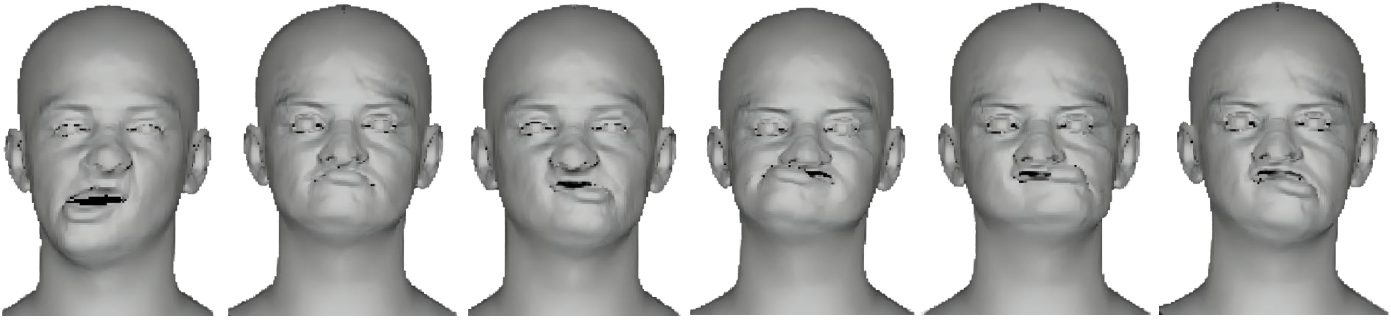}
  \caption{Visualizations of failure cases.}
  \label{failure_cases}
\end{figure}

We present several visualizations of failure cases here.  
Although most test cases produce plausible results, we still observe a few 
abnormal ones, as illustrated in Figure~\ref{failure_cases}. In particular, the model 
occasionally generates exaggerated jaw movements that deviate from natural 
facial expressions. Interestingly, such anomalies usually occur in only a 
few isolated frames within an coherent sequence.  

We speculate that these errors are likely caused by rare but large 
inter-frame variations in the training data. 
Despite thorough manual inspection, annotators may inevitably overlook some anomalous frames during video playback, reflecting the inherent limitations of human visual perception.
For future work, we plan to explore anomaly detection or interpolation 
strategies to smooth these irregular frames. 
Alternatively, we may incorporate automatic detection of abnormal 
expressions during data preprocessing and apply interpolation to 
mitigate dataset noise.   

\end{document}